\documentclass[runningheads]{llncs}
\usepackage{graphicx}

\usepackage{xcolor,booktabs}         
\usepackage{adjustbox,multirow}

\begin{document}

\title{AutoSAM: Adapting SAM to Medical Images by Overloading the Prompt Encoder}

\author{Tal Shaharabany \and
Aviad Dahan \and
Raja Giryes \and Lior Wolf}
\authorrunning{Shaharabany et al.}
\titlerunning{AutoSAM: Adapting SAM to Medical Images}
\institute{Tel-Aviv University}

\maketitle

\begin{abstract}
The recently introduced Segment Anything Model (SAM) combines a clever architecture and large quantities of training data to obtain remarkable image segmentation capabilities. However, it fails to reproduce such results for Out-Of-Distribution (OOD) domains such as medical images. Moreover, while SAM is conditioned on either a mask or a set of points, it may be desirable to have a fully automatic solution. In this work, we replace SAM's conditioning with an encoder that operates on the same input image. By adding this encoder and without further fine-tuning SAM, we obtain state-of-the-art results on multiple medical images and video benchmarks. This new encoder is trained via gradients provided by a frozen SAM. For inspecting the knowledge within it, and providing a lightweight segmentation solution, we also learn to decode it into a mask by a shallow deconvolution network.  
\end{abstract}

\section{Introduction}

\begin{figure}[t]
    \centering
    \begin{tabular}{ccc}
    \includegraphics[width=0.3023413\linewidth]{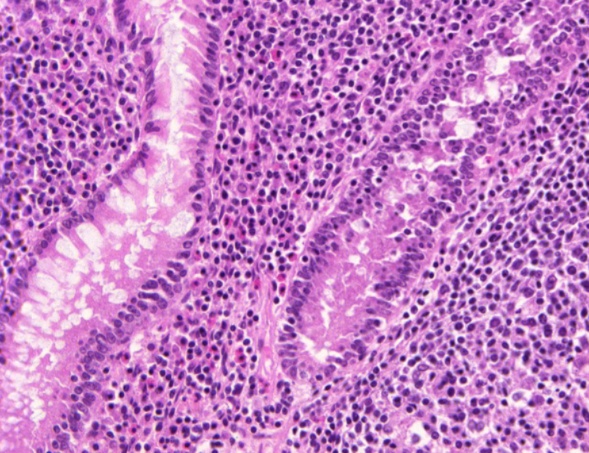} &
    \includegraphics[width=0.3023413\linewidth]{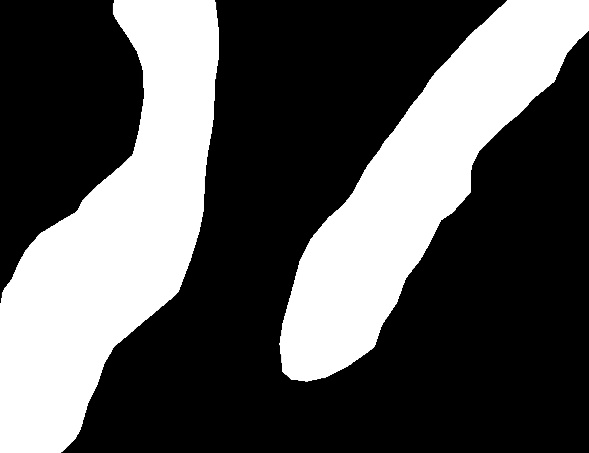} &
    \includegraphics[width=0.3023413\linewidth]{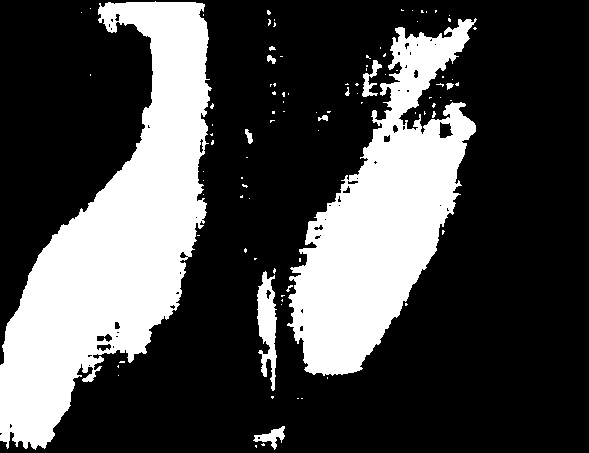} \\
    (a) & (b) & (c) \\
    \includegraphics[width=0.3023413\linewidth]{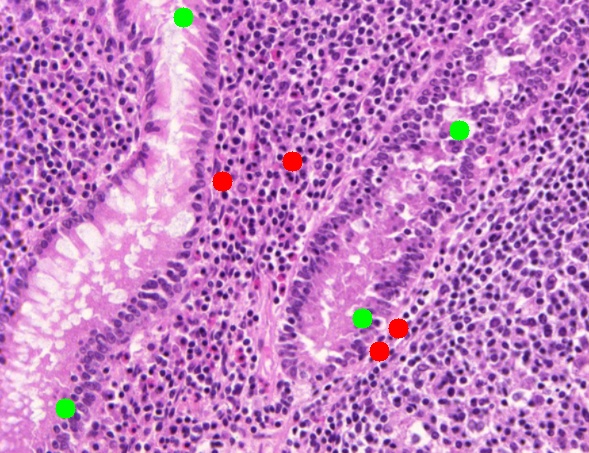} &
    \includegraphics[width=0.3023413\linewidth]{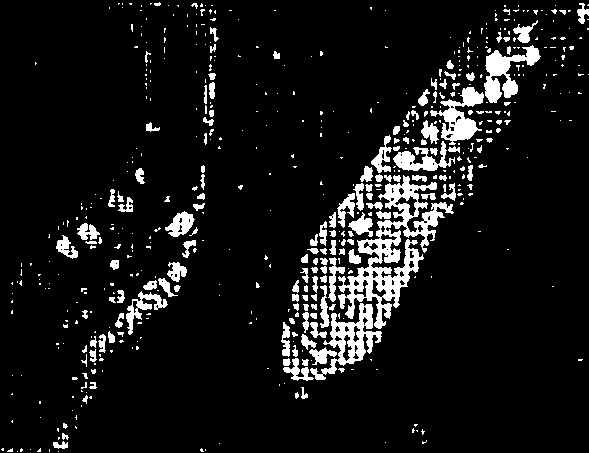} &
    \includegraphics[width=0.3023413\linewidth]{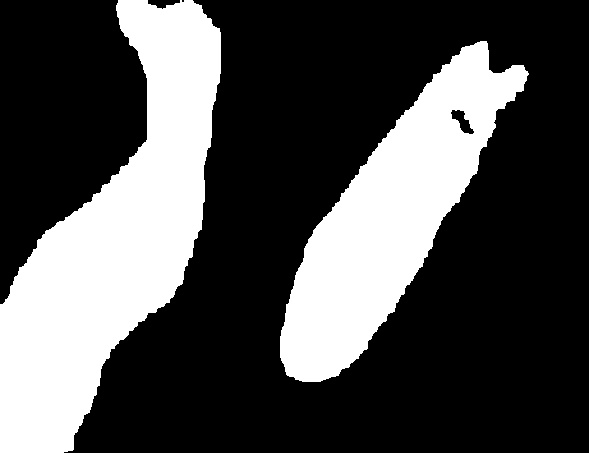} \\
    (d) & (e) & (f) \\
    \end{tabular}
    \caption{An example of segmenting an image from the Glas dataset. (a) the input image. (b) the ground truth mask. (c) the results of SAM with the GT mask provided to its mask encoder. (d) a point-based prompt. (e) SAM's result based on the point prompt. (f) our result, where the input image itself is given as a prompt to the prompt-encoder we train.} 

    \label{fig:teaser}
\end{figure}

The promptable image segmentation model, SAM~\cite{kirillov2023segment}, is an efficient and practical approach to real-world segmentation tasks that allows for flexibility in prompts, quick mask computation, and ambiguity awareness. However, SAM's performance may not be optimal on medical imaging datasets due to its pre-training on natural images as illustrated in Fig.~\ref{fig:teaser}.

In this paper, we propose an end-to-end approach to improve segmentation mask accuracy for medical images without fine-tuning the pretrained SAM network. Our solution involves the training of an auxiliary prompt encoder network, which generates a surrogate prompt for SAM given an input image. While the prompt encoder provided with SAM can accept inputs such as a bounding box, a set of points, or a mask, the one we train has the image itself as its input. 
We term this overloading, since in object-oriented programming, overloading is a feature that allows a class to have multiple methods with the same name, but with different types of input parameters. 

During training, the SAM network propagates gradients to the prompt encoder network from a binary cross-entropy loss and a Dice loss. The encoder network that we train employs the Harmonic Dense Net \cite{chao2019hardnet} as its backbone and has significantly fewer learnable parameters than SAM's own decoders. 
As mentioned, the main SAM network is not modified, which makes our method easy to implement and avoids finding a suitable training schedule for SAM fine-tuning.

We have evaluated our method on multiple publicly available medical images and videos datasets. Our results show a significant improvement in segmentation performance compared to the baseline method and other state-of-the-art approaches.

\section{Related Work}
Medical image segmentation is an active research area that plays a vital role in diagnosis~\cite{devunooru2021deep}, treatment planning~\cite{sharma2010automated}, and disease monitoring~\cite{norman2018use}. U-net~\cite{ronneberger2015u} has been widely used for various medical image segmentation tasks. Over the years, various modifications and versions have been proposed for the U-net segmentation architecture~\cite{zhou2018unet++,xiao2018weighted,wang2021uctransnet,patel2021enhanced,shaharabany2022end}.

Our solution is based on SAM~\cite{kirillov2023segment}, which is based on a visual-transformer~\cite{strudel2021segmenter}, similar to other segmentation architectures~\cite{strudel2021segmenter}. SAM~\cite{kirillov2023segment} is trained on the largest segmentation dataset reported to date, comprising over 1 billion masks on 11 million licensed and privacy-respecting natural images. The model serves as an effective foundation model and its zero-shot performance is comparable to or better than many fully supervised results in natural image segmentation. Moreover, its modular and promptable design enables transfer learning to new tasks and image distributions. In this work, we harness these properties in order to achieve SOTA results on out-of-distribution (OOD) data by replacing SAM's built-in prompt encoder with our custom encoder.

Concurrent with our work, Zu et al.~\cite{wu2023medical}
 fine-tune SAM's encoder and decoder using adaptation blocks (this technique is used as a baseline in~\cite{hu2021lora}). The prompt encoders of SAM are not tuned and this method, therefore, requires a prompt in the form of positive points. In our method, we replace the prompt encoder.  The encoder that we train receives the same input image as the main network, hence the name AutoSAM. Note that we do not fine-tune the encoder and decoder of SAM. Moreover, as we show in Section~\ref{sec:experiments}, our encoder can be easily converted to a segmentation network by simply adding a few convolutional layers to it and training them for this task.

In modern Large Language Models~\cite{touvron2023llama} and promptable text-image models such as diffusion models~\cite{ramesh2021zero,saharia2022photorealistic}, a careful prompt can draw the line between a desired outcome and an unusable result. The task of learning the desired prompt for a specific outcome from the model without training its weights can be achieved using various strategies such as prompt-engineering~\cite{shin2020autoprompt} and prompt learning~\cite{wallace2019universal,li2021prefix,gao2020making,liu2021gpt,tewel2021zero,tewel2022zeroshot}. Our work utilizes a Pseudo-Token optimization method for learning the optimal prompt embeddings OOD samples.

\section{Method}

\begin{figure}[t]
    \centering
    \includegraphics[width=1.01\linewidth]{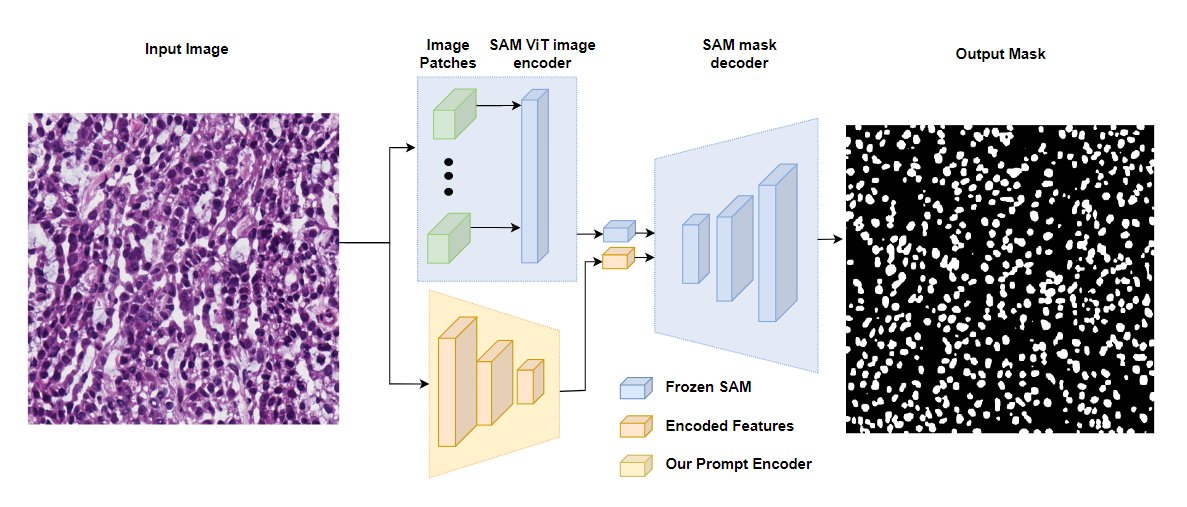}
    \caption{An illustration of AutoSAM. SAM's prompt encoder is replaced with our custom encoder while the image encoder and mask decoder are frozen.}
    \label{fig:Arch}
\end{figure}

SAM, the promptable image segmentation model, is built to be efficient and practical for real-world use. To support flexibility in prompts, quick mask computation, and ambiguity awareness, SAM is designed with three components.

First, a robust image encoder $E_s$ computes an image embedding for an input image $I$. Second, a prompt encoder $E_M$ embeds prompts for use in the segmentation process. Lastly, a lightweight mask decoder $D_s$ predicts segmentation masks based on the combined information from the image and prompt encoders.

SAM's design allows for the reuse of the same image embedding with different prompts, thereby achieving efficient computation. This separation of components is crucial to enable SAM to support a wide range of prompts and perform computation in real-time.

Since SAM is trained on over 1 billion masks from 11 million natural images, its performance on medical imaging datasets may not be optimal. We present an end-to-end approach to improve segmentation mask accuracy in this domain, without fine-tuning the pretrained SAM network, as presented in Fig.~\ref{fig:Arch}.

The SAM network $S$ produces an output segmentation mask $M_z$ by taking the input image $I$ and the prompts' embedding $Z$:

\begin{equation}
\label{eq:sam}
   M_z = S(I,Z),
\end{equation}
The prompts embedding $Z$ can be any representation of different prompts, such as masks, boxes, and points.

Instead of using the original prompts encoder, we introduce a prompts generator network, denoted as $g$, that generates guidance prompts $Z_I$ for SAM given an input image $I$. $g$ is the only network trained by our method.

This prompts generator network $g$ takes as input the image $I$ and generates prompts $Z_I=g(I)$ for SAM to improve its segmentation mask output.

While training our method, the SAM network $S$ propagates gradients to the prompts generator network $g$ from two segmentation losses that we employ: the binary cross-entropy loss (BCE) and the Dice loss.  The BCE loss is given by the negative log-likelihood of the ground truth mask $M$ and the SAM output $S(I,Z_I)$, while the Dice loss measures the overlap between the predicted and ground truth masks. Formally, the losses are expressed as:

\begin{equation}
\label{eq:g}
L_{seg}(I) = L_{BCE}(I,Z_I,M) + L_{dice}(I,Z_I,M),
\end{equation}
where the BCE loss is defined as:

\begin{equation}
\label{eq:bce}
L_{BCE}(I,Z,M) = - M*log(S(I,Z)) - (1 - M)*log(1 - S(I,Z))\,.
\end{equation}
The Dice loss is defined as:
\begin{equation}
\label{eq:dice}
\mathcal{L}_{dice}(I,Z,M) = 1 - \frac{2TP(S(I,Z),M) + 1}{2TP(S(I,Z),M) + FN(S(I,Z),M) + FP(S(I,Z),M) + 1 }\,,
\end{equation}
where TP, FN, and FP denote the true positive, false negative, and false positive, respectively, between the ground truth mask $M$ and the output mask $S(I,Z)$.
To simplify the implementation, we do not use weighting for the loss terms.

\paragraph{Architecture}
The proposed architecture for $g$ employs the Harmonic Dense Net~\cite{chao2019hardnet} as its backbone. This network comprises six ``HarD'' blocks, each with output channels of 192, 256, 320, 480, 720, and 1280, respectively. We initialize the network with pretrained ImageNet weights.

The decoder of $g$ includes two upsampling blocks that produce a resolution of $64 \times 64$ with 256 output channels. Each block consists of two convolutional layers with a kernel size of 3 and zero padding of one. Additionally, we apply batch normalization after the last convolution layer and before the activation function. The activation function of the first layer is ReLU, while the second layer uses $tanh$. Each layer receives a skip connection from the encoder block with the same spatial resolution. Notably, our decoder requires significantly fewer learnable parameters than a regular decoder, and fewer skip connections are used in the encoder since only two blocks are employed there.

{\color{black} In terms of FLOPs, our model uses 25.11 GMACs for an image size of $256^2$, whereas SAM uses 2733.31 GMACs for an image size of $1024^2$ (fixed size of the ViT) only for the image encoder. The peak memory consumption is 371MB for our model and 6006MB for SAM image encoder. Therefore, the overhead of our encoder is almost negligible. The number of parameters of $g$ is 41.56M while SAM ViT has 637M.}

\paragraph{A surrogate decoder for $g(I)$} To gain insight into the information provided by the encoder we train, we decode $g(I)$ as a mask. For this purpose, we learn a mapping $h$ from the space of encoded images $g(I)$ to the corresponding ground truth mask $M$. 

This surrogate decoder $h$ minimizes a segmentation loss very similar to Eq.~\ref{eq:g}, except that it compares $h(g(I))$ with $M$, for a fixed $g$. The architecture of $h$ comprises two deconvolution layers that produce a map with a resolution of $256 \times 256$, making it a lightweight alternative to SAM.

As it turns out, despite its size, $h(g(I))$ is often a reasonable segmentation mask, see Sec.~\ref{sec:experiments}. However, it is not as powerful as AutoSAM, which applies SAM to $g(I)$. 

\begin{figure}[htb!]
\setlength{\tabcolsep}{3.5pt} 
    \centering
    \begin{tabular}{ccccc}
    \includegraphics[width=0.1652384\linewidth]{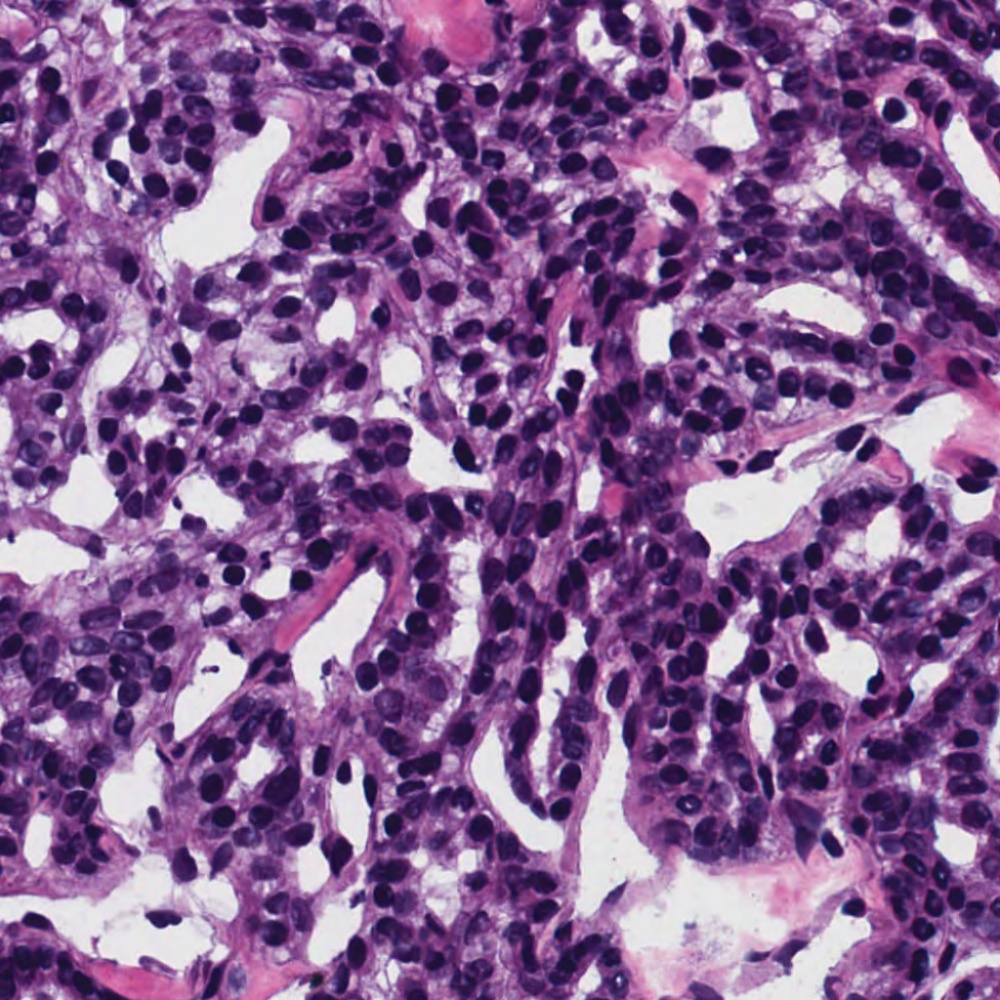} &
    \includegraphics[width=0.1652384\linewidth]{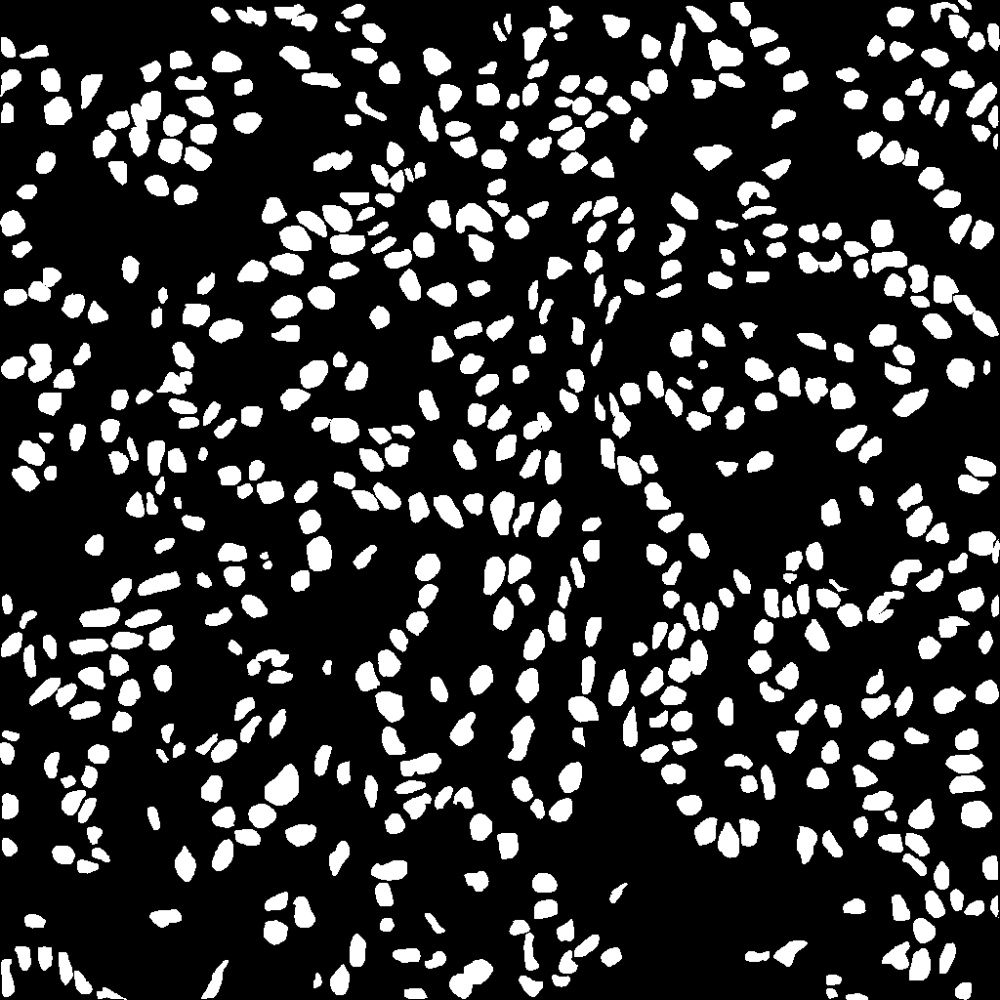} &
    \includegraphics[width=0.1652384\linewidth]{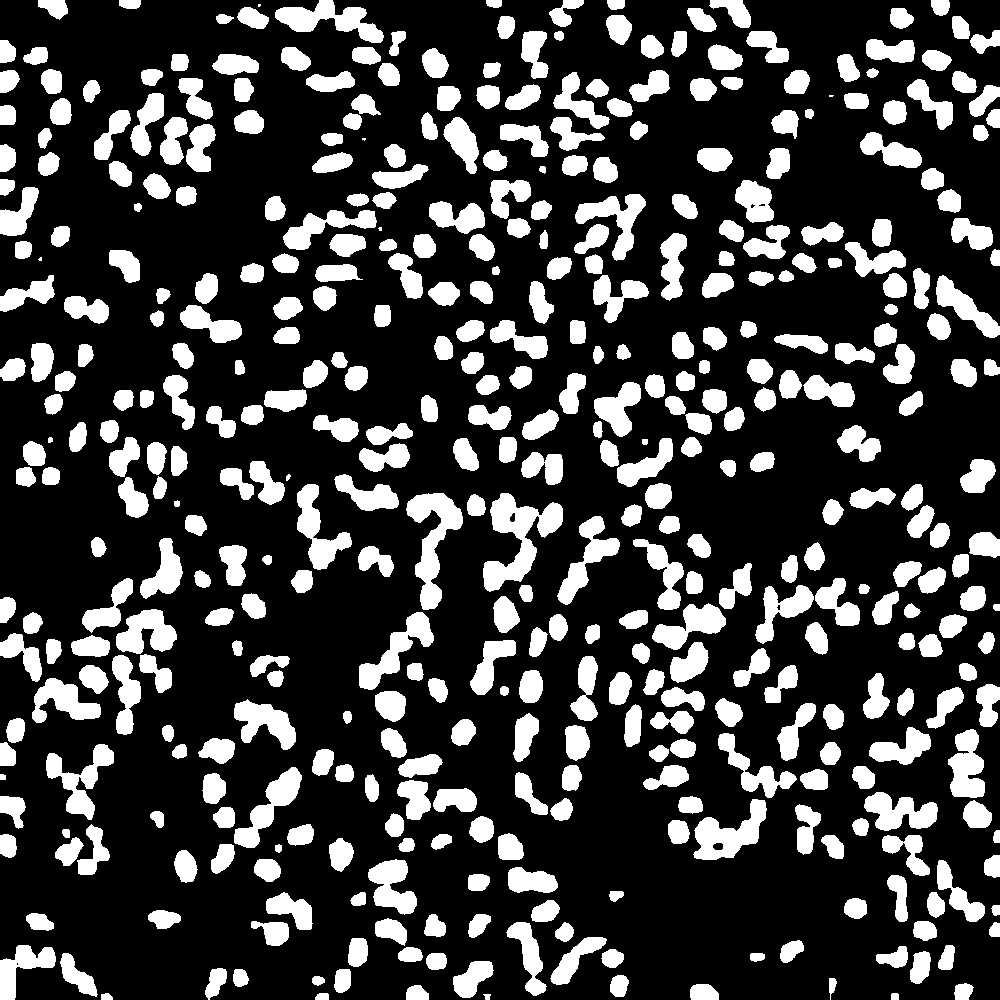} &
    \includegraphics[width=0.1652384\linewidth]{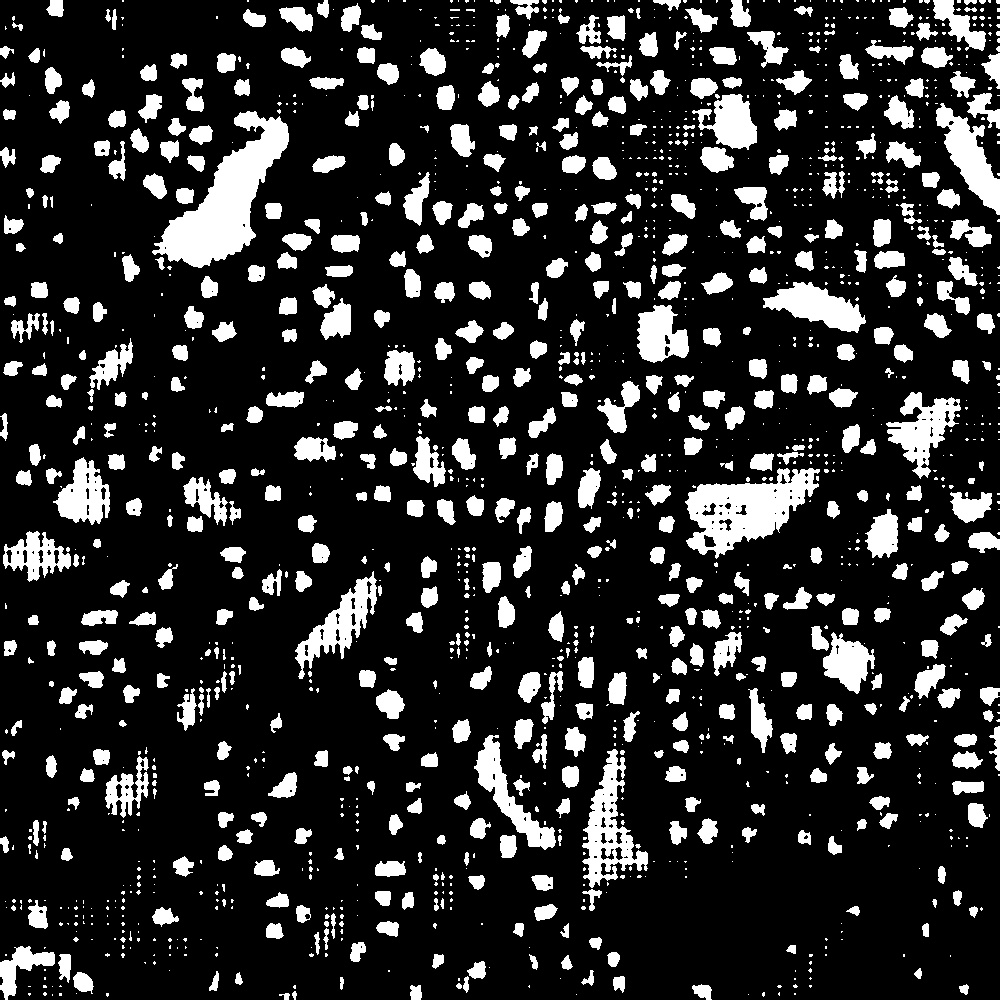} &
    \includegraphics[width=0.1652384\linewidth]{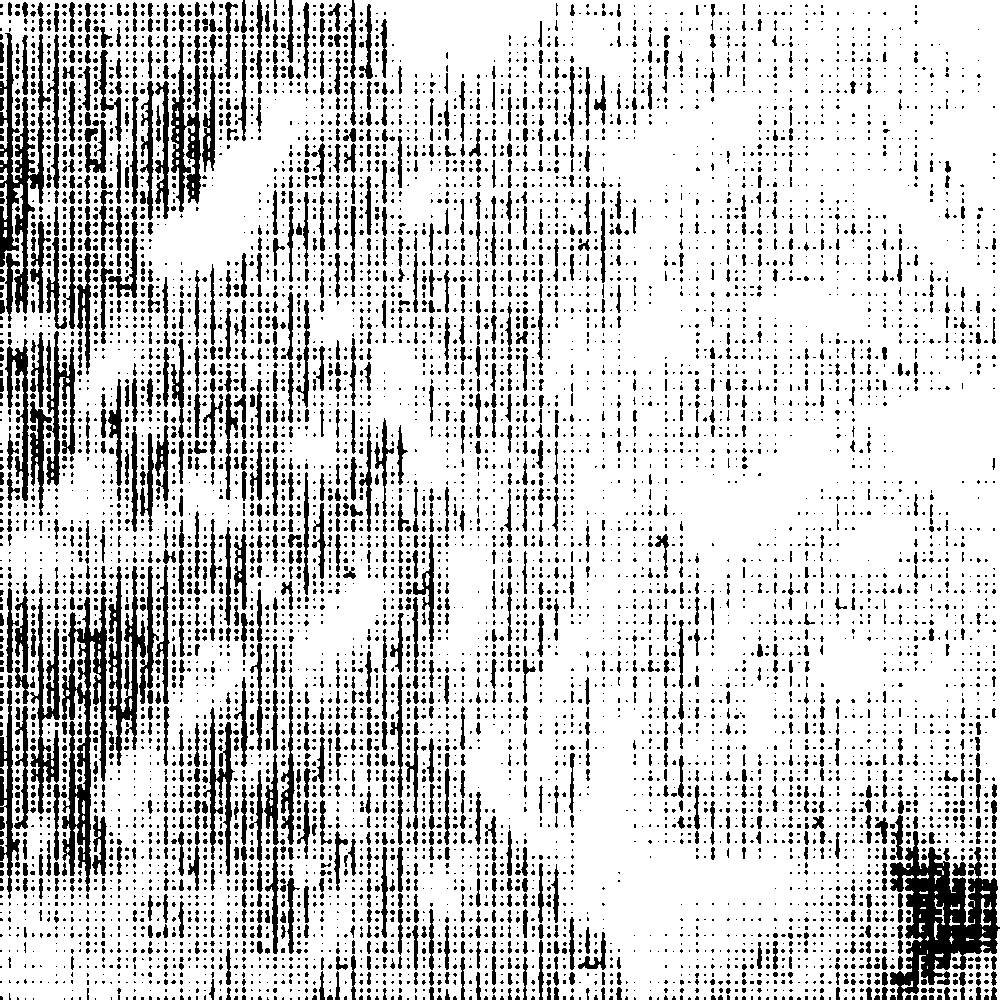} \\

    \includegraphics[width=0.1652384\linewidth]{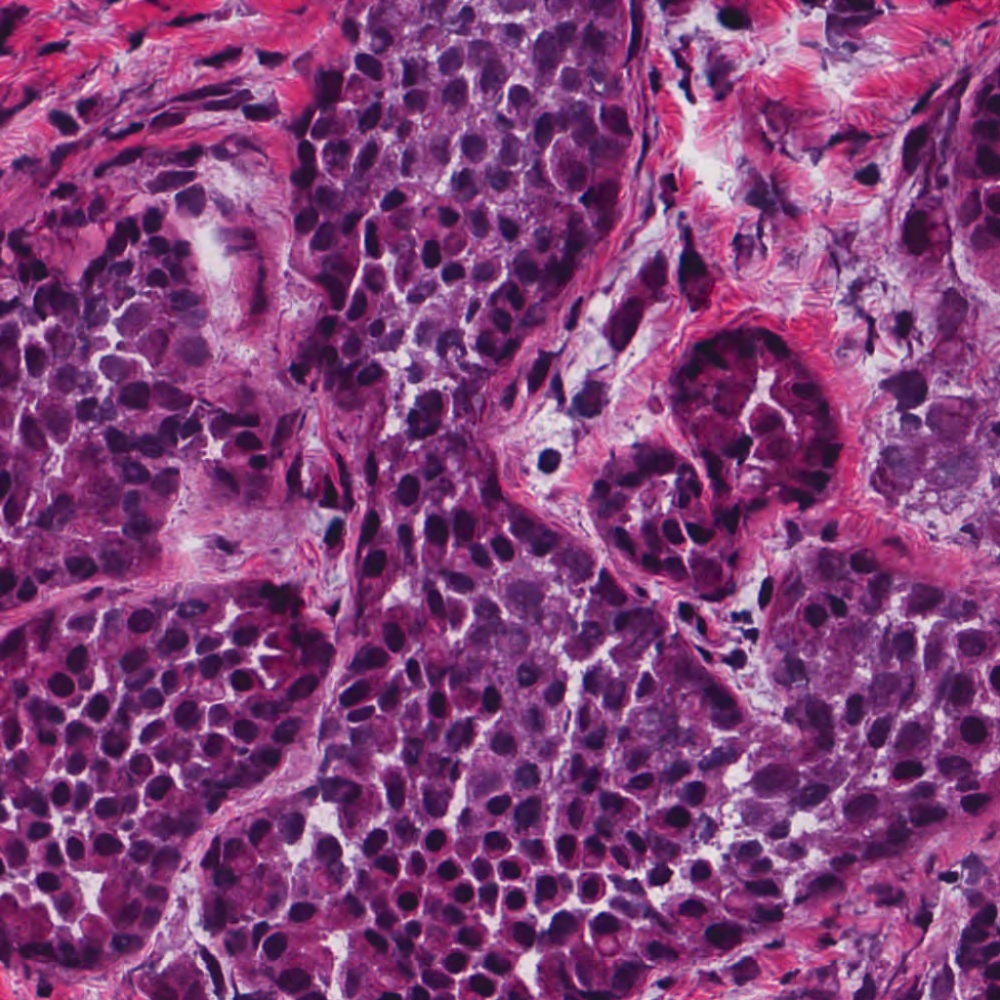} &
    \includegraphics[width=0.1652384\linewidth]{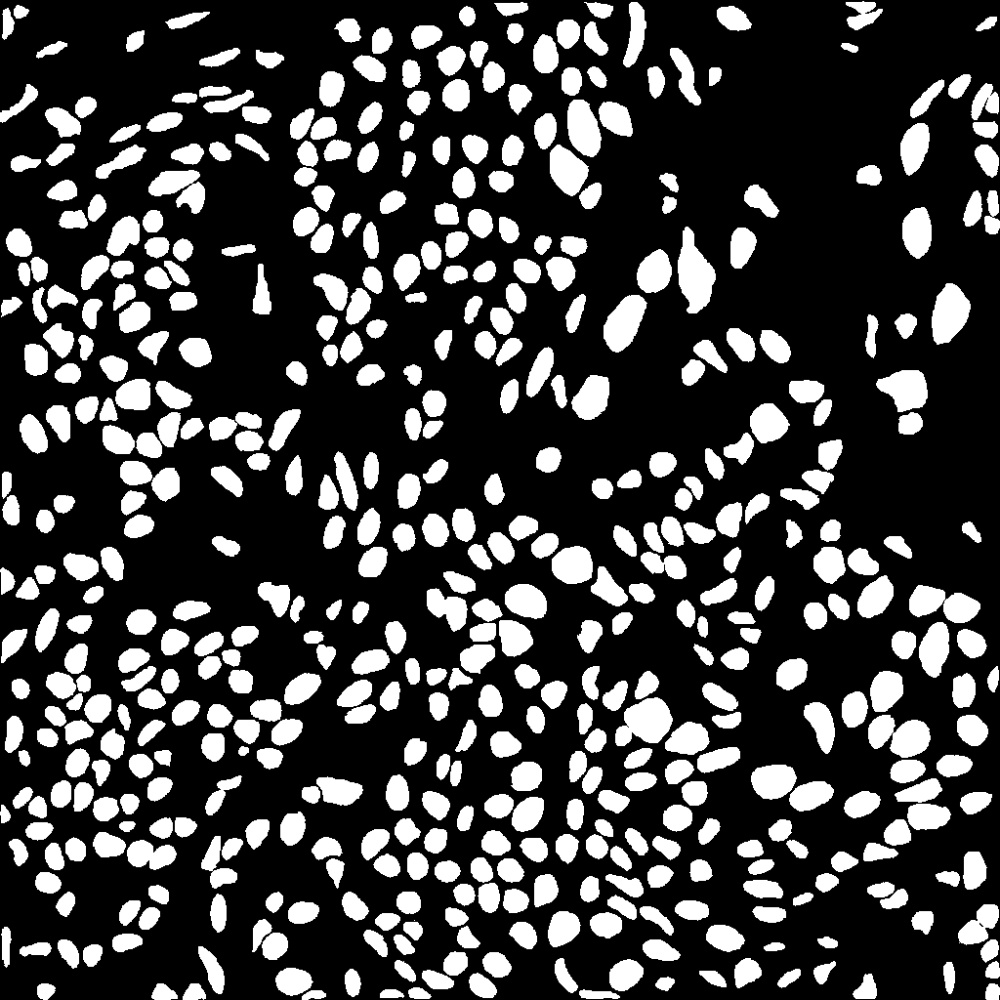} &
    \includegraphics[width=0.1652384\linewidth]{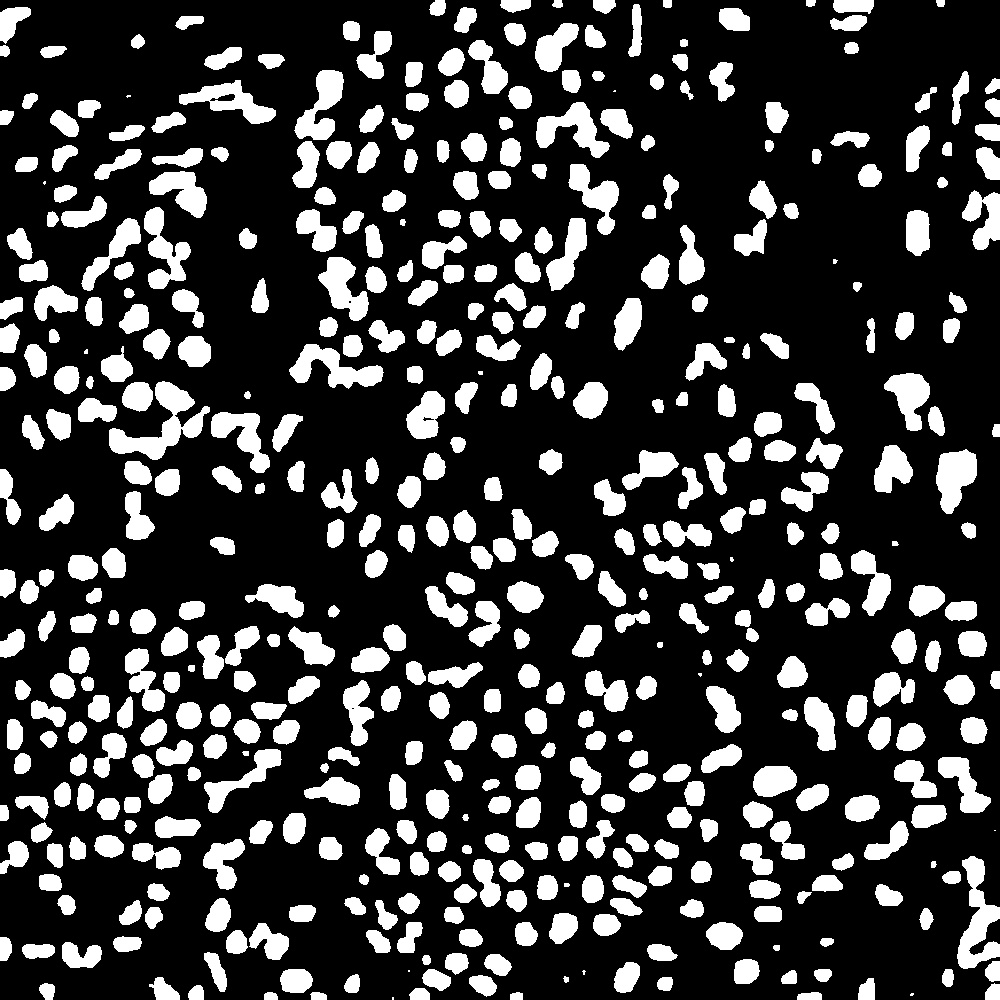} &
    \includegraphics[width=0.1652384\linewidth]{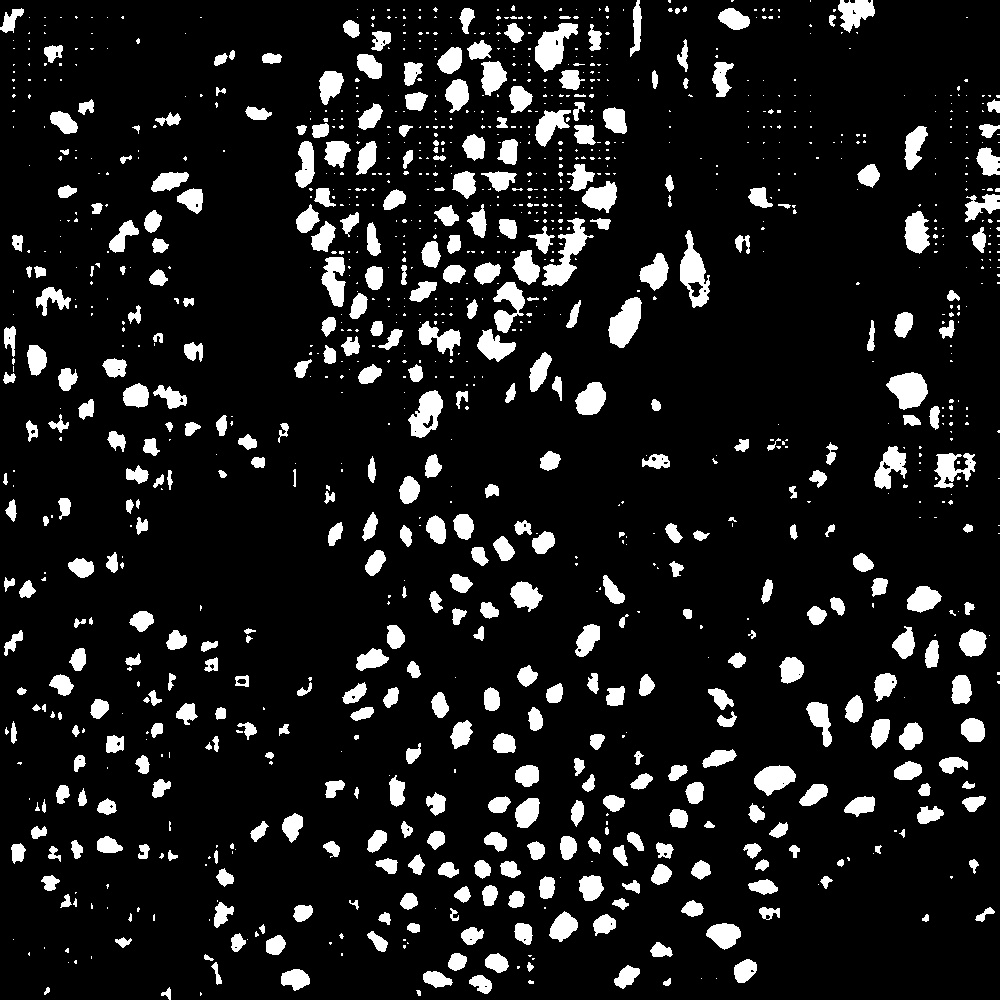} &
    \includegraphics[width=0.1652384\linewidth]{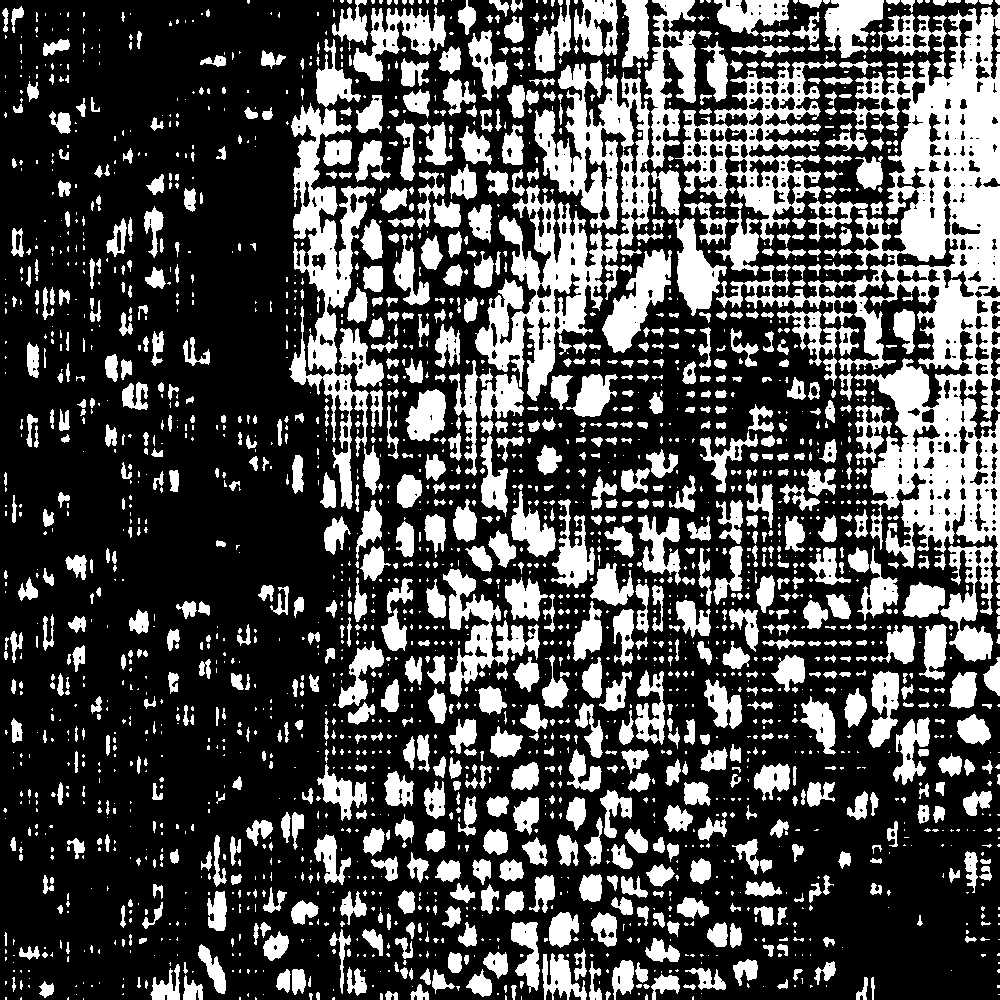} \\
    
    \includegraphics[width=0.1652384\linewidth]{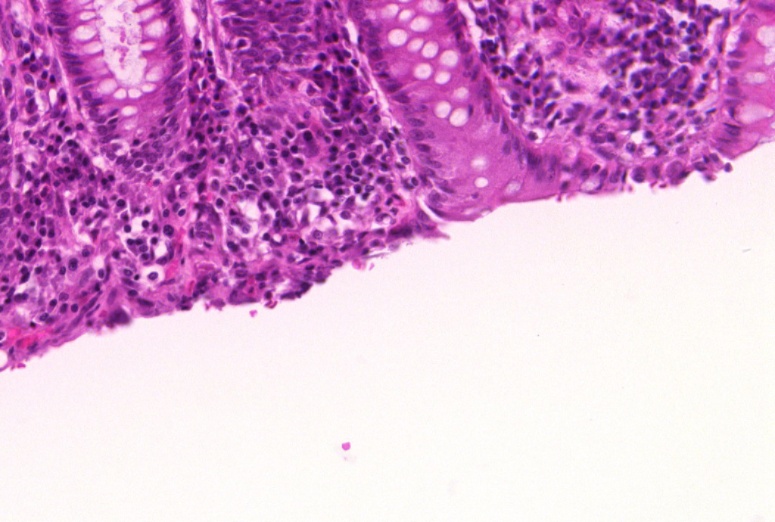} &
    \includegraphics[width=0.1652384\linewidth]{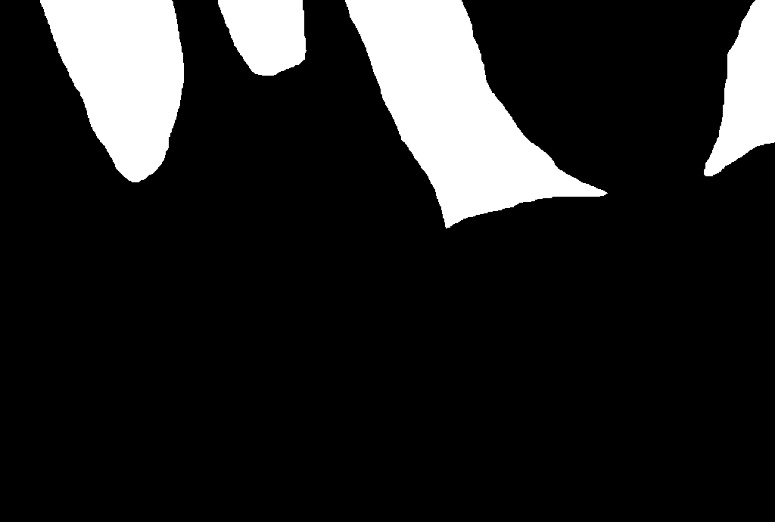} &
    \includegraphics[width=0.1652384\linewidth]{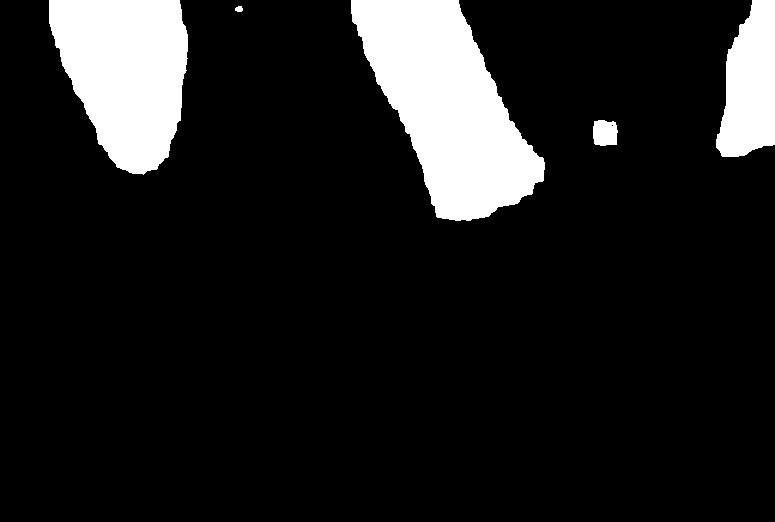} &
    \includegraphics[width=0.1652384\linewidth]{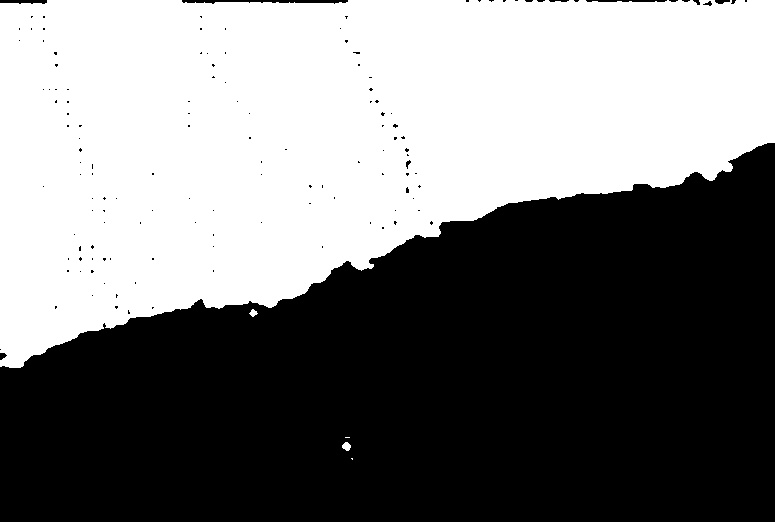} &
    \includegraphics[width=0.1652384\linewidth]{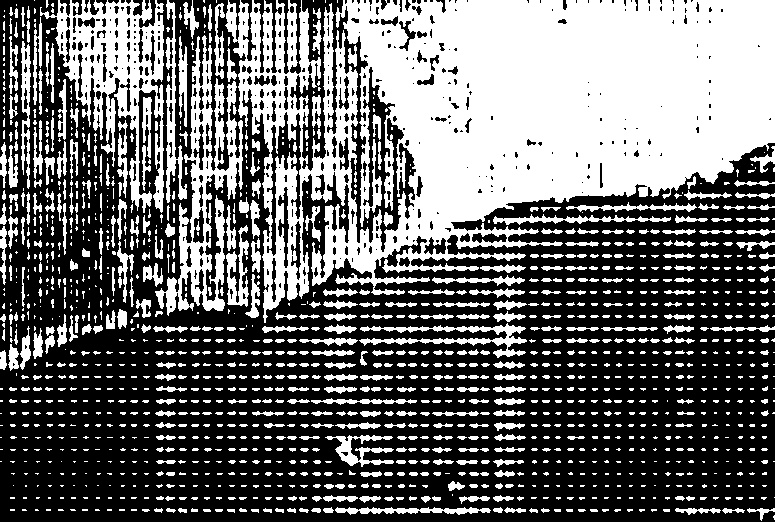} \\

     \includegraphics[width=0.1652384\linewidth]{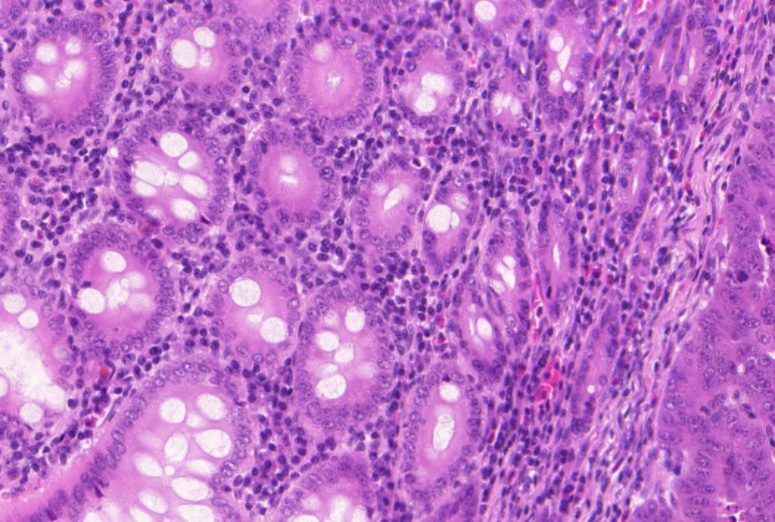} &
    \includegraphics[width=0.1652384\linewidth]{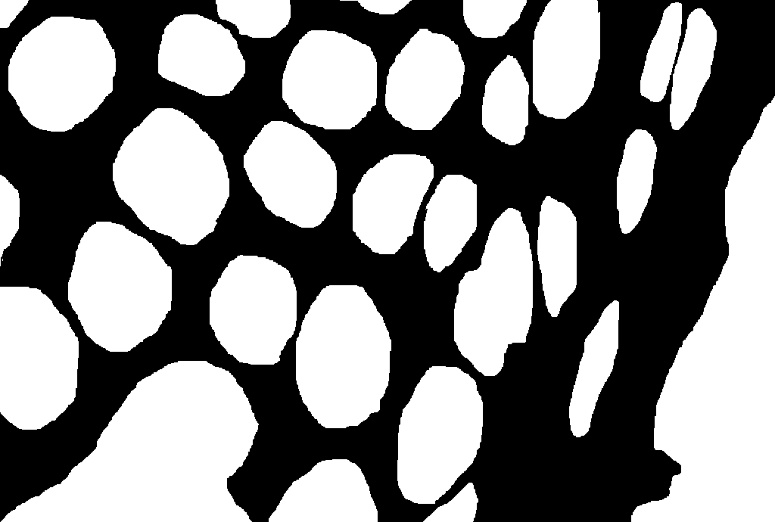} &
    \includegraphics[width=0.1652384\linewidth]{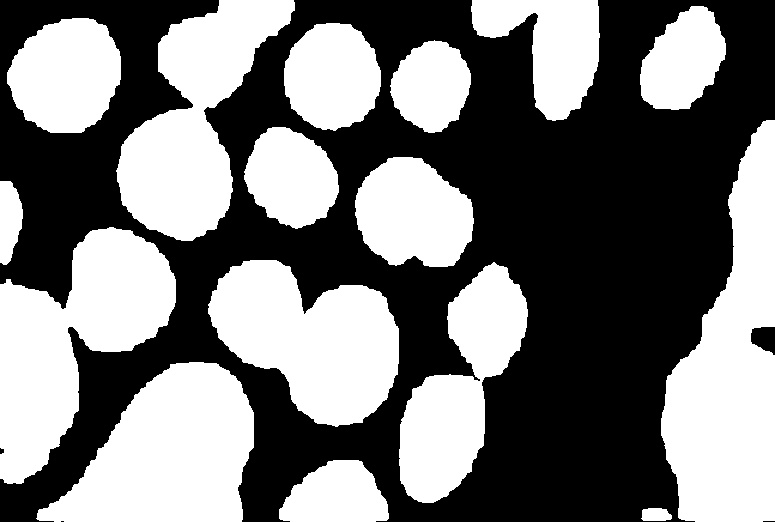} &
    \includegraphics[width=0.1652384\linewidth]{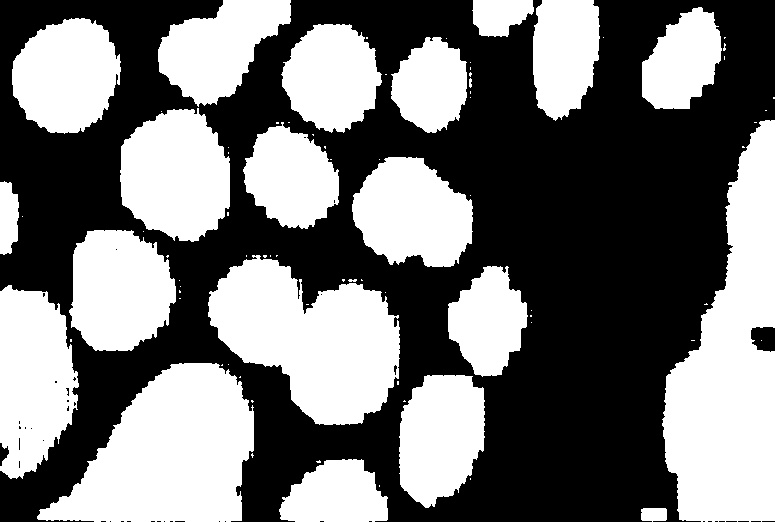} &
    \includegraphics[width=0.1652384\linewidth]{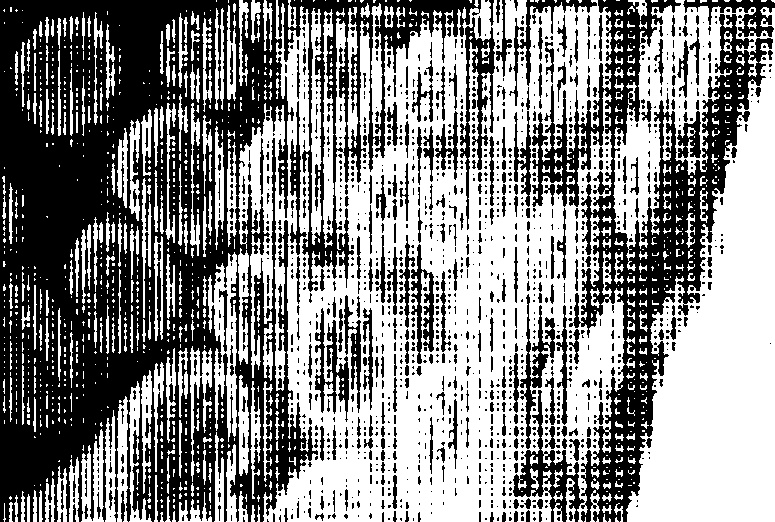} \\
    
    \includegraphics[width=0.1652384\linewidth]{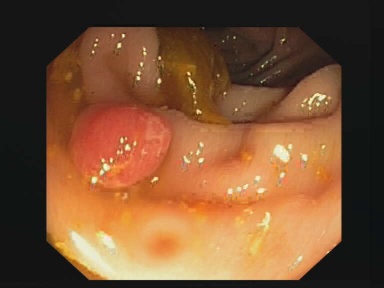} &
    \includegraphics[width=0.1652384\linewidth]{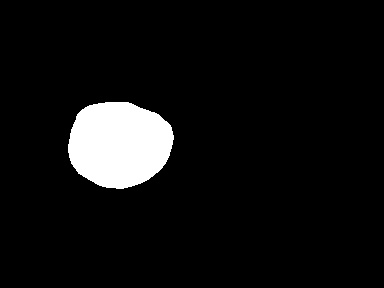} &
    \includegraphics[width=0.1652384\linewidth]{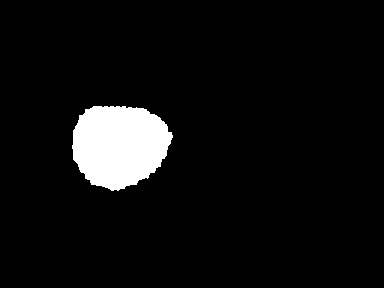} &
    \includegraphics[width=0.1652384\linewidth]{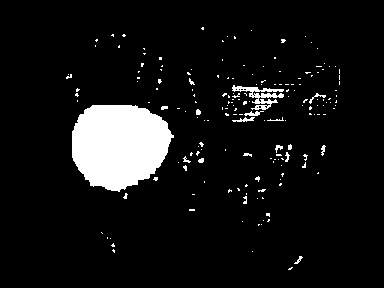} &
    \includegraphics[width=0.1652384\linewidth]{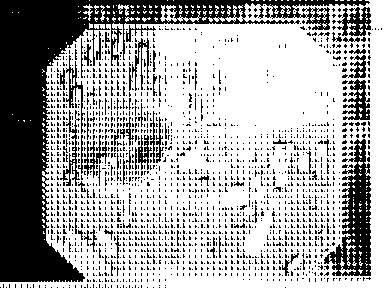} \\

    \includegraphics[width=0.1652384\linewidth]{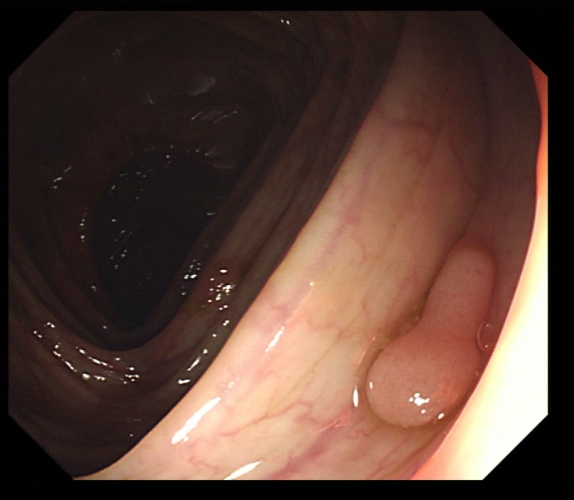} &
    \includegraphics[width=0.1652384\linewidth]{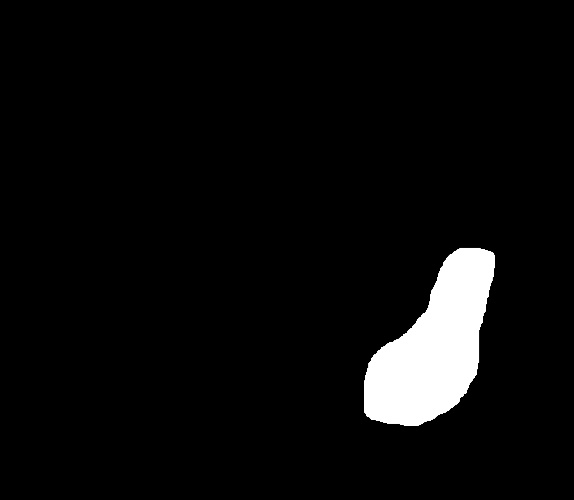} &
    \includegraphics[width=0.1652384\linewidth]{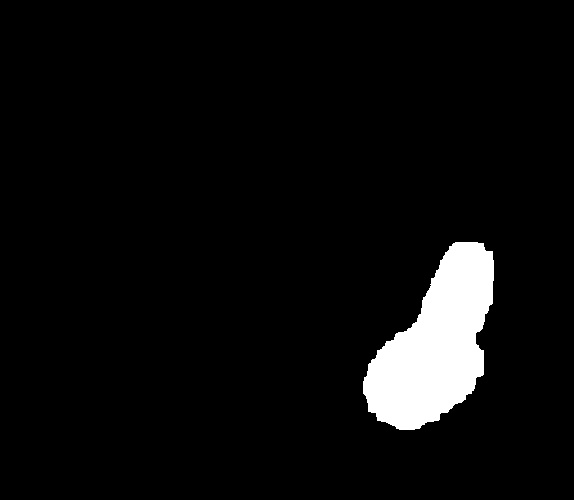} &
    \includegraphics[width=0.1652384\linewidth]{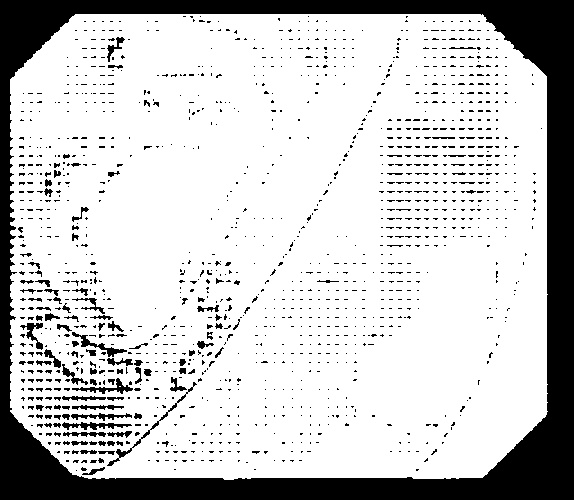} &
    \includegraphics[width=0.1652384\linewidth]{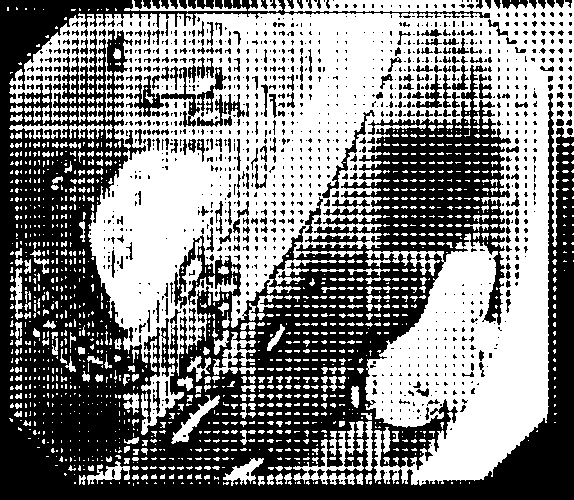} \\

    (a) & (b) & (c) & (d) & (e) 
    \end{tabular}
    \caption{Sample results of the proposed method on the Nucleus challenges (MoNuSeg) - rows 1,2. The gland segmentation dataset (Glas) rows 3,4. The Kvasir polyp segmentation dataset rows 5,6 where (a) Input image. (b) Ground truth segmentation. (c) The final segmentation map $M_z$. (d) output of SAM with our mask as input to the mask prompt encoder. (e) output of SAM with the ground truth mask as input to the same prompt encoder.}
    \label{fig:Monu_seg}
\end{figure}

\section{Experiments}
\label{sec:experiments}

In this study, we evaluate our proposed method on multiple medical datasets. We compare our results with state-of-the-art methods and present a number of exloratory results. 

\paragraph{Datasets}
The MoNuSeg dataset~\cite{kumar2019multi} comprises 30 microscopic images from seven organs in the training set, with annotations of 21,623 individual nuclei, and 14 similar images in the test set. To be consistent with previous work, we resize the images to $512\times 512$\cite{valanarasu2021medical} and employ an encoder-decoder architecture based on the HarDNet-85\cite{chao2019hardnet} backbone. 

The Gland segmentation (GlaS) challenge~\cite{sirinukunwattana2017gland} comprises 85 images for training and 80 for testing, with all images resized to $224\times 224$ following \cite{wang2021uctransnet}. 

We also evaluated our algorithm on four Polyp datasets: Kvasir-SEG~\cite{jha2020kvasir}, ClinicDB~\cite{bernal2015wm}, ColonDB~\cite{tajbakhsh2015automated}, and ETIS~\cite{silva2014toward}, following \cite{fan2020pranet}. We split the data into a training set of 1448 images, comprising 900 images from ClinicDB and 548 from Kvasir, and a test set comprising 100 images from ClinicDB, 64 from Kvasir, 196 from ETIS, and 380 from ColonDB. 

Lastly, our method was tested on the SUN-SEG Video-Polyp-Segmentation database, based on~\cite{misawa2020development,ji2022video}. The colonoscopy videos are from Showa University and Nagoya University database (also named SUN-database)~\cite{misawa2020development}.
The initial classification information and bounding box annotations are provided by three research assistants and examined by two expert endoscopists with professional domain knowledge. The SUN dataset is then extended by Ji et al.~\cite{ji2022video} to have various annotations such as object masks, boundaries, scribbles, and polygons.
The original SUN database has $113$ colonoscopy videos, including $100$ positive cases with $49,136$ polyp frames and $13$ negative cases with $109,554$ non-polyp frames.
in their work Ji et al.~\cite{ji2022video} manually trim them into $378$ positive and $728$ negative clips while maintaining their consecutive intrinsic relationship.
Such data preprocessing ensures that each clip has around 3-11s duration at a real-time frame rate (i.e., 30 fps), promoting the fault-tolerant margin for various algorithms and devices.
Overall, the SUN-SEG database contains $1,106$ short video clips with $158,690$ video frames total.
Although being a video-segmentation task, we have chosen to use our architecture without any modification, using a single frame at a time as the input without relying on temporal data whatsoever. This image-based architecture achieved SOTA performance in almost every metric, competing with video-based methods as shown in table~\ref{tab:video}.

\paragraph{Training details} During the training of our network, we employ the ADAM optimizer with an initial learning rate of $0.0003$, and a weight decay regularization parameter set to $1\cdot10^{-5}$. A batch size of $10$ is utilized, and we conduct training on NVIDIA A6000 with 48GB GPU RAM. The maximum number of epochs for network training was set to 200. The SAM pre-trained weights that we utilized were based on the ViT `huge' architecture. SAM received an input image size of $1024 \times 1024$ as per the original algorithm.

To ensure fairness in comparison with the state-of-the-art method 3P-SEG~\cite{shaharabany2022end}, we employed identical data augmentations during training. For the GlaS dataset, we applied a set of augmentations that included: (i) color jitter with the parameters of brightness sampled uniformly between $[0,0.2]$, contrast in the range $[0,0.2]$, saturation in the range $[0,0.2]$, and hue in the range $[0,0.1]$; (ii) a random horizontal flip; and (iii) a random affine transformation with a translation of 5 and scale of $(0,0.2)$. For the MoNu dataset, we utilized (i) a random rotation augmentation of \textpm20 degrees and a scale range of $[0.75, 1.25]$; (ii) a random horizontal flip with a probability of 0.5; and (iii) random color jitter with a maximal value of 0.4 for brightness, 0.4 for contrast, 0.4 for saturation, and 0.1 for hue.

During the training of the lightweight decoder $h$, we utilized the ADAM optimizer with an initial learning rate of 0.0003 and set the weight decay regularization parameter to $1\cdot10^{-5}$. We trained with a batch size of 24 on an NVIDIA A5000 with 24GB GPU RAM and set the maximum number of iterations for network training to 60.

\paragraph{Evaluation Metrics} For evaluating the performance of our network on image-based segmentation tasks, we employed the widely-used evaluation metrics of Mean Intersection-over-Union (IoU) and Dice-Score. Specifically, we computed the IoU by dividing the area of overlap between the ground truth (GT) masks and the network's output mask by the area of union between the two masks. Moreover, we computed the Dice-Score as a measure of the overlap between the two masks, by taking twice the area of overlap and dividing it by the sum of the areas of the two masks. Both metrics were computed after thresholding the network's output mask to obtain a binary mask that separates the foreground and background regions. 

As for the video segmentation task, following~\cite{ji2022video}, we use six different metrics for model evaluation between prediction $P_s$ and ground-truth $G_s$ at timestamp $s$. These metrics are as follows:
(a) Dice coefficient (${\rm Dice} = \frac{ 2 \times \vert P_s \cap G_s \vert}{\vert P_s \cup G_s \vert}$).
The operators $\cap$, $\cup$, and $\vert \cdot \vert$ denote the intersection, union, and the number of pixels in an area, respectively.
(b) Pixel-wise sensitivity (${\rm Sen} = \frac{ \vert P_s \cap G_s \vert}{\vert G_s \vert}$).
(c) F-measure~\cite{achanta2009frequency}. The harmonic mean of precision and recall, weighted by $\beta$, ($F_{\beta} = \frac{(1+\beta^{2}) \times {\rm Prc} \times {\rm Rcl}}{\beta^{2} \times ( {\rm Prc} + {\rm Rcl} )}$).
This metric is widely used in measuring binary masks by combining both precision (${\rm Prc} = \frac{\vert P_{s} \cap G_{s} \vert}{\vert P_{s}\vert}$) and recall (${\rm Rcl} = \frac{\vert P_{s} \cap G_{s} \vert}{\vert G_{s}\vert}$) for more comprehensive evaluation.
(d) Weighted F-measure~\cite{margolin2014evaluate} ($F_{\beta}^{w} = \frac{(1+\beta^{2}) \times {\rm Prc}^{w} \times {\rm Rcl}^{w}}{\beta^{2} \times ( {\rm Prc}^{w} + {\rm Rcl}^{w} )}$). This metric, suggested by~\cite{fan2021cognitive,cheng2021structure} amends the ``Equal importance flaw'' in Dice and $F_{\beta}$, providing more reliable evaluation results. As for  $\beta^{2}$, we set this factor of $F_{\beta}$ and $F_{\beta}^{w}$ to be 0.3 and 1, respectively, following~\cite{ji2022video,borji2015salient}.
(e) Structure measure~\cite{fan2017structure} ($\mathcal{S}_{\alpha} = \alpha \times \mathcal{S}_{o}(P_{s}, G_{s}) + (1-\alpha) \times \mathcal{S}_{r}(P_{s}, G_{s})$).
This metric is used to measure the structural similarity at object-aware $\mathcal{S}_{o}$ and region-aware $\mathcal{S}_{r}$, respectively.
we set $\alpha$ = 0.5.
(f) Enhanced-alignment measure, proposed by~\cite{fan2018enhanced} is a human visual perception-based metric,: $E_{\phi} = \frac{1}{W \times H} \sum_{x}^{W} \sum_{y}^{H} \phi (P_{s}(x,y), G_{s}(x,y))$, where $\phi$ is the enhanced-alignment matrix.
$W$ and $H$ are the width and height of ground-truth $G_{s}$.

\begin{figure}[t!]
    \setlength{\tabcolsep}{2.5pt} 
    \renewcommand{\arraystretch}{1} 
    \centering
    \begin{tabular}{ccccc}
    \includegraphics[width=0.1588942384\linewidth]{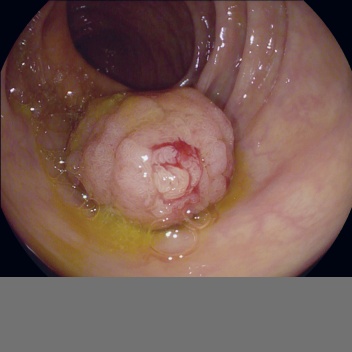} &
    \includegraphics[width=0.1588942384\linewidth]{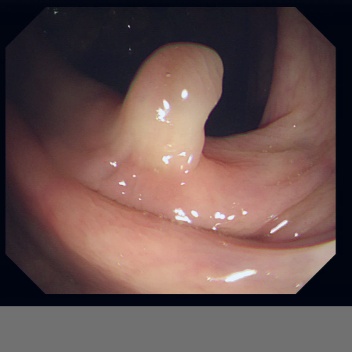} &
    \includegraphics[width=0.1588942384\linewidth]{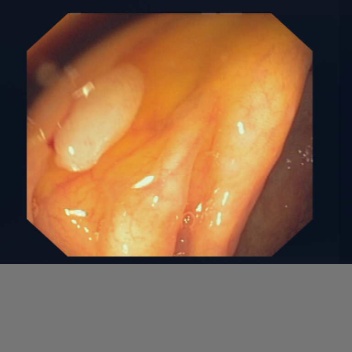} &
    \includegraphics[width=0.1588942384\linewidth]{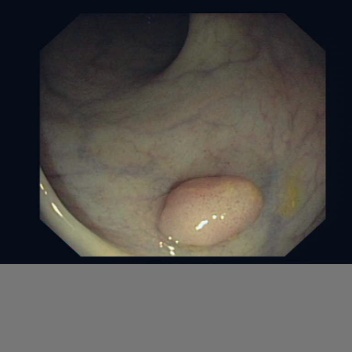} &
    \includegraphics[width=0.1588942384\linewidth]{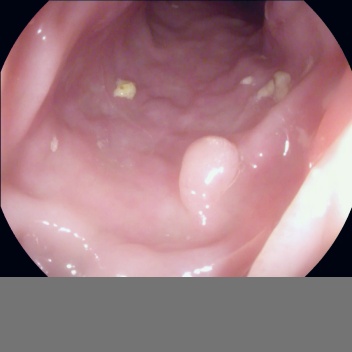} \\
    \includegraphics[width=0.1588942384\linewidth]{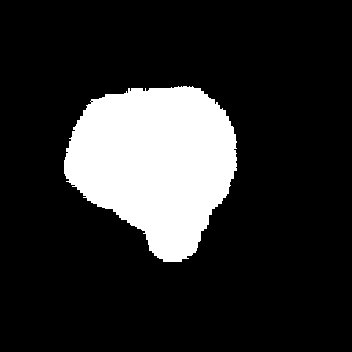} &
    \includegraphics[width=0.1588942384\linewidth]{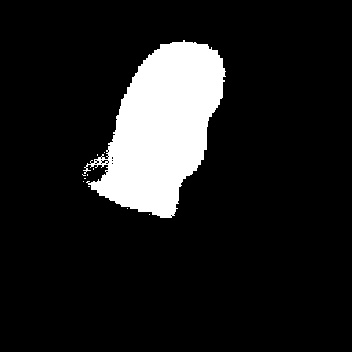} &
    \includegraphics[width=0.1588942384\linewidth]{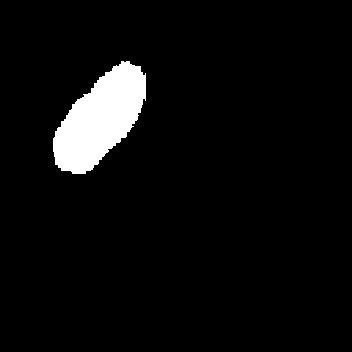} &
    \includegraphics[width=0.1588942384\linewidth]{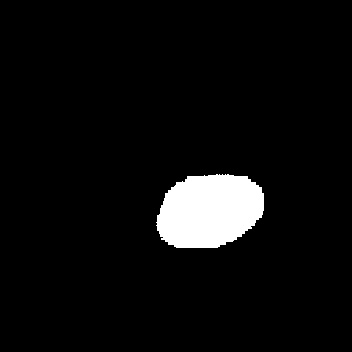} &
    \includegraphics[width=0.1588942384\linewidth]{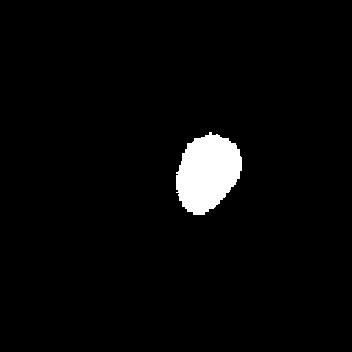} \\
\includegraphics[width=0.1588942384\linewidth]{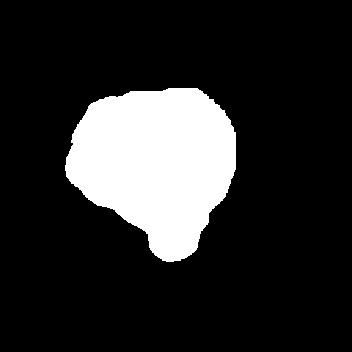} &
    \includegraphics[width=0.1588942384\linewidth]{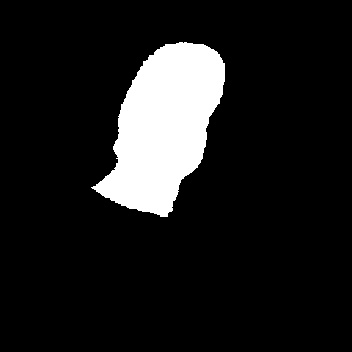} &
    \includegraphics[width=0.1588942384\linewidth]{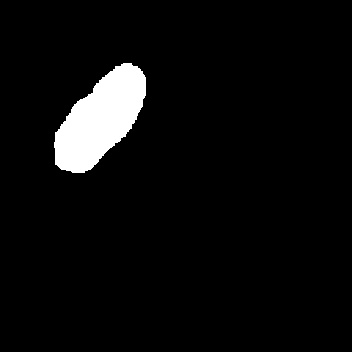} &
    \includegraphics[width=0.1588942384\linewidth]{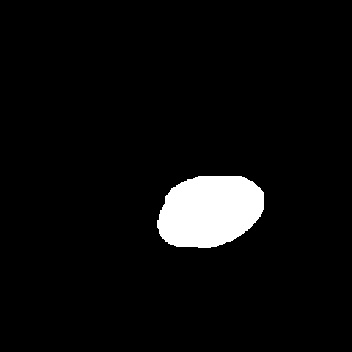} &
    \includegraphics[width=0.1588942384\linewidth]{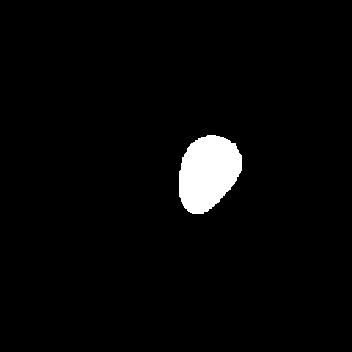} \\
    \includegraphics[width=0.1588942384\linewidth]{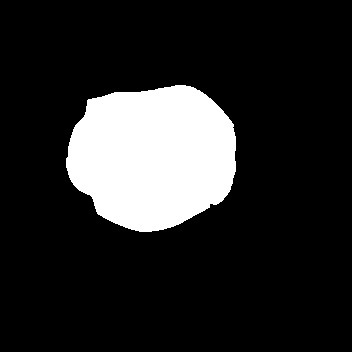} &
    \includegraphics[width=0.1588942384\linewidth]{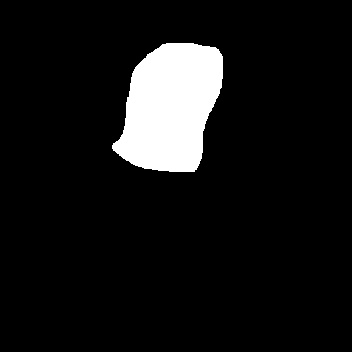} &
    \includegraphics[width=0.1588942384\linewidth]{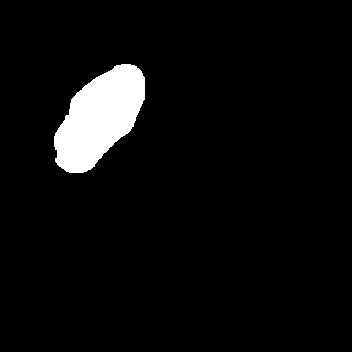} &
    \includegraphics[width=0.1588942384\linewidth]{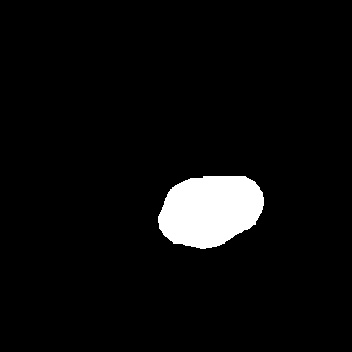} &
    \includegraphics[width=0.1588942384\linewidth]{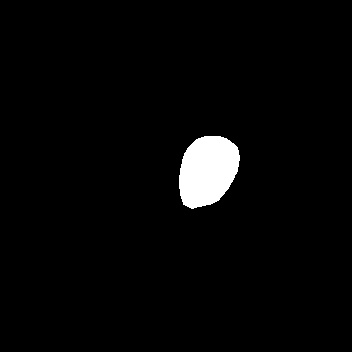} \\
    \end{tabular}
    \caption{The results of the lightweight decoder $h$ on sample test images. The first row shows the input image $I$, the second row shows $h(g(I))$, which is the segmentation mask obtained with the surrogate decoder $h$, the third depicts the results of AutoSAM using the same $g(I)$, and the last row shows the ground-truth segmentation mask $M$.}
    \label{fig:inverting}
\end{figure}

\begin{figure}[b!]
\setlength{\tabcolsep}{3.5pt} 
    \renewcommand{\arraystretch}{2} 
    \centering
    \begin{tabular}{cccc}
    \includegraphics[width=0.2242384\linewidth]{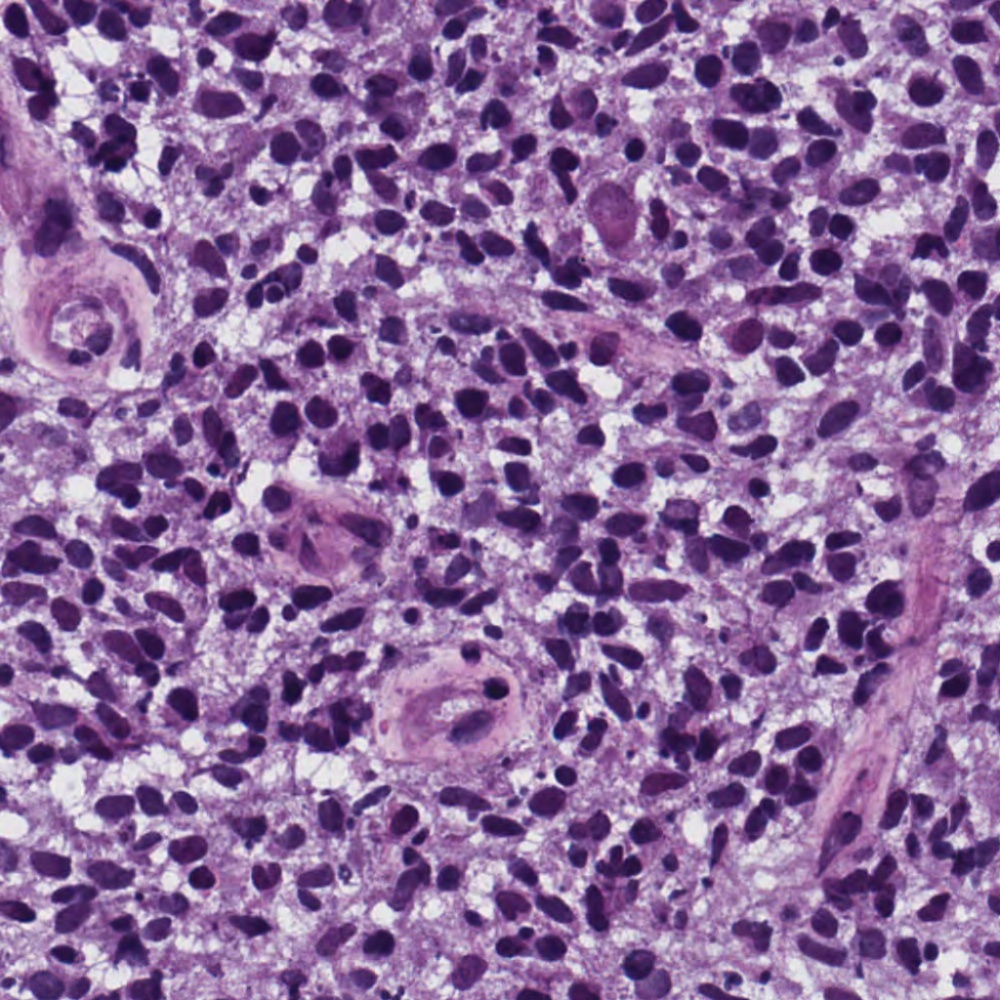} &
    \includegraphics[width=0.2242384\linewidth]{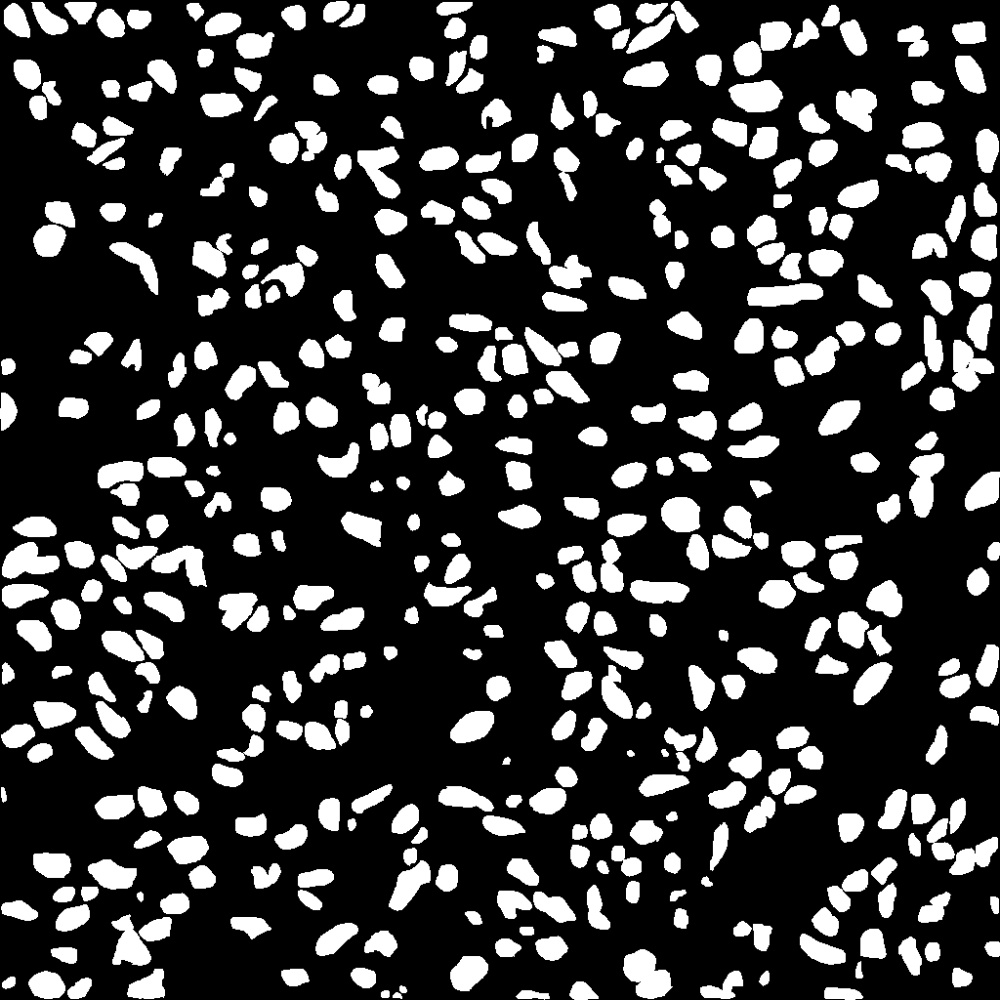} &
    \includegraphics[width=0.2242384\linewidth]{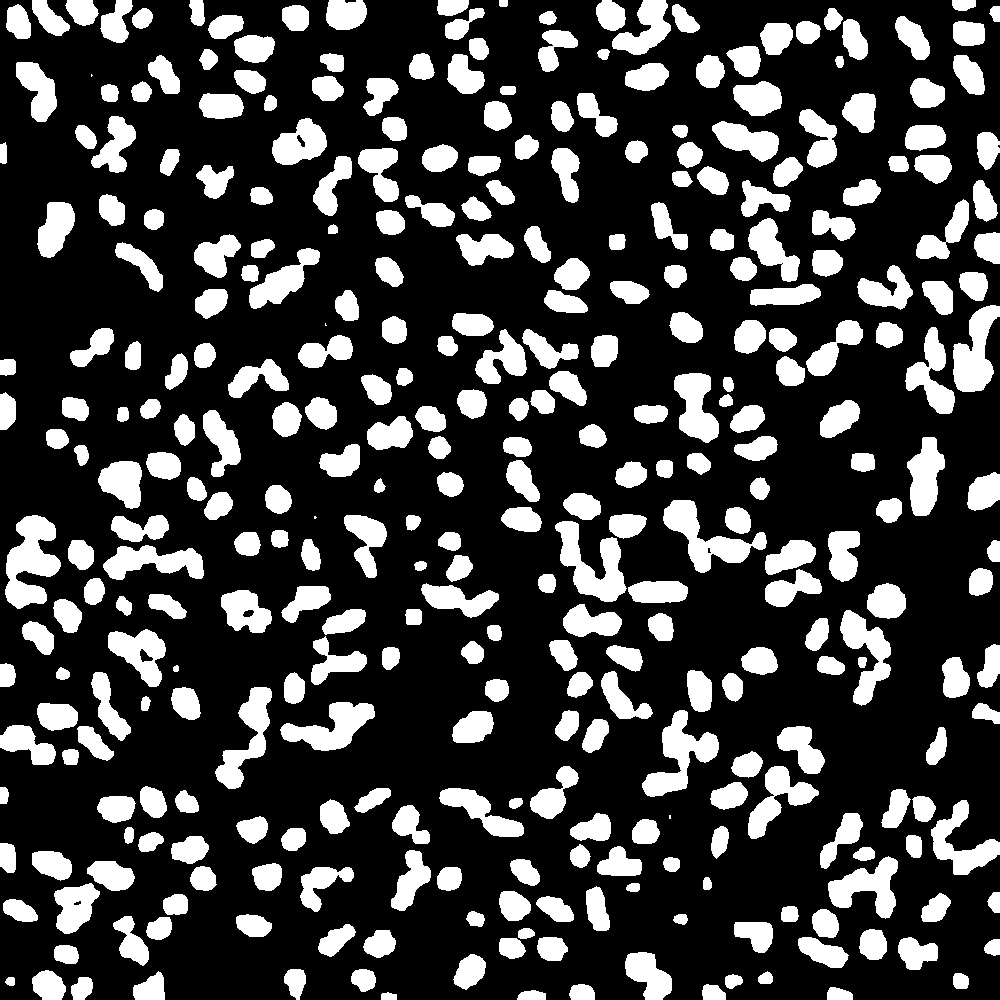} &
    \includegraphics[width=0.2242384\linewidth]{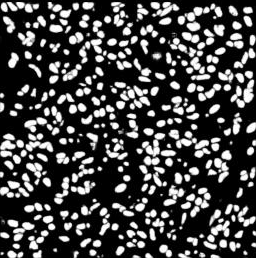} \\

    \includegraphics[width=0.2242384\linewidth]{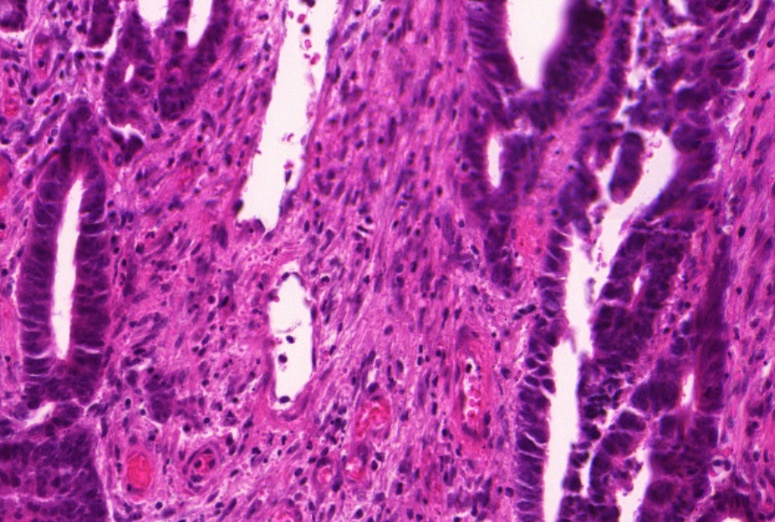} &
    \includegraphics[width=0.2242384\linewidth]{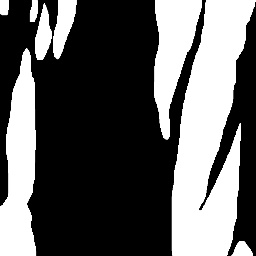} &
    \includegraphics[width=0.2242384\linewidth]{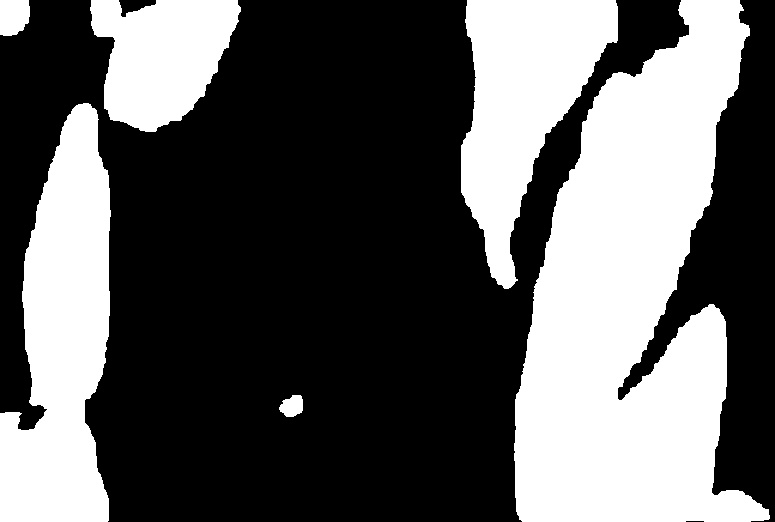} &
    \includegraphics[width=0.2242384\linewidth]{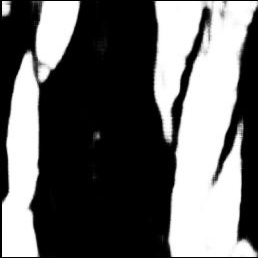} \\

    (a) & (b) & (c) & (d) 
    \end{tabular}
    \caption{A visual comparison of our solution to MedAdapterSAM~\cite{wu2023medical} for Glas and Monu datasets, where (a) input image (b) ground-truth mask (c) our solution (d) MedAdapterSAM~\cite{wu2023medical} output.}
    \label{fig:MedAdapter}
\end{figure}

\begin{table}[b!]
\begin{center}
    \begin{tabular}{lcccc}
    \toprule
    \multirow{2}{*}{Method}&\multicolumn{2}{c}{Monu}&\multicolumn{2}{c}{GlaS}\\ \cmidrule(rl){2-3}
    \cmidrule(rl){4-5}
     & Dice & IoU &  Dice & IoU \\
    \midrule
    FCN~\cite{badrinarayanan2017segnet} & 28.84 &  28.71 & - & - \\
    U-Net~\cite{ronneberger2015u} & 79.43 & 65.99 & 86.05 & 75.12\\
    U-Net++~\cite{zhou2018unet++} & 79.49 & 66.04 & 87.36 & 79.03\\
    Res-UNet~\cite{xiao2018weighted} & 79.49 & 66.07 & - & - \\
    Axial Attention~\cite{wang2020axial} & 76.83 & 62.49 & - & - \\
    MedT~\cite{valanarasu2021medical} & 79.55 & 66.17 & 88.85 & 78.93\\
    FCN-Hardnet85~\cite{chao2019hardnet} & 79.52 & 66.06 & 89.37 & 82.09\\
    UCTransNet~\cite{wang2021uctransnet} & 79.87 & 66.68 & 89.84 & 82.24\\
    3P-SEG~\cite{shaharabany2022end} & 80.30 & 67.19 & 91.19 & 84.34\\
    MedAdaptor-SAM~\cite{wu2023medical} (conditioned on GT points) & 80.34 & 67.33 & 92.02 & 85.88\\
    AutoSAM (ours)& \textbf{82.43} & \textbf{70.17}  & \textbf{92.82} & \textbf{87.08}\\
    \midrule
    Lightweight decoder $h(g(I))$ & 76.75 & 62.32 & 91.51 & 84.80\\
    SAM w/ GT point prompt & 29.65 & 17.52 & 61.67 & 46.40\\
    SAM w/ GT mask as prompt & 30.24 & 18.21 & 58.46 & 42.81\\
    SAM w/ AutoSAM output as the mask prompt & 58.10 & 41.26 & 87.71 & 79.92 \\
    \bottomrule
    \end{tabular}
    \end{center}
    \caption{MoNu and GlaS results. Our method achieves SOTA results on both datasets. MedAdaptor-SAM requires point input as a prompt. GT=ground truth.}
    \label{tab:Monu}
    \end{table}

\begin{table}[t]
\begin{center}
    \begin{tabular}{@{}l@{~}cccccccc@{}}
    \toprule
    \multirow{2}{*}{Method}&\multicolumn{2}{c}{Kvasir33~\cite{jha2020kvasir}}&\multicolumn{2}{c}{Clinic~\cite{bernal2015wm}}&\multicolumn{2}{c}{Colon~\cite{tajbakhsh2015automated}}&\multicolumn{2}{c}{ETIS~\cite{silva2014toward}}\\ \cmidrule(l){2-9} 
     & Dice & IoU &  Dice & IoU  & Dice & IoU &  Dice & IoU \\
    \midrule
    U-Net~\cite{ronneberger2015u} & 81.8 & 74.6 & 82.3 & 75.5 & 51.2 & 44.4 & 39.8 & 33.5\\
    U-Net++~\cite{zhou2018unet++} & 82.1 & 74.3 & 79.4 & 72.9 & 48.3 & 41.0 & 40.1 & 34.4\\
    SFA~\cite{fang2019selective} & 72.3 & 61.1 & 70.0 & 60.7 & 46.9 & 34.7 & 29.7 & 21.7\\
    MSEG~\cite{huang2021hardnet} & 89.7 & 83.9 & 90.9 & 86.4 & 73.5 & 66.6 & 70.0 & 63.0\\
    DCRNet~\cite{yin2021duplex} & 88.6 & 82.5 & 89.6 & 84.4 & 70.4 & 63.1 & 55.6 & 49.6\\
    ACSNet~\cite{zhang2020adaptive} & 89.8 & 83.8 & 88.2 & 82.6 & 71.6 & 64.9 & 57.8 & 50.9\\
    PraNet~\cite{fan2020pranet} & 89.8 & 84.0 & 89.9 & 84.9 & 71.2 & 64.0 & 62.8 & 56.7\\
    EU-Net~\cite{patel2021enhanced} & 90.8 & 85.4 & 90.2 & 84.6 & 75.6 & 68.1 & 68.7 & 60.9\\
    SANet~\cite{wei2021shallow} & 90.4 & 84.7 & 91.6 & 85.9 & 75.3 & 67.0 & 75.0 & 65.4\\
    Polyp-PVT~\cite{dong2021polyp} & 91.7 & 86.4 & 93.7 & 88.9 & 80.8 & 72.7 & 78.7 & 70.6\\
    FCN-Hardnet85~\cite{chao2019hardnet} & 90.0 & 84.9 & 92.0 & 86.9 & 77.3 & 70.2 & 76.9 & 69.5\\
    3P-SEG~\cite{shaharabany2022end} & \textbf{91.8} & {86.5} & \textbf{93.8} & {89.0} & {80.9} & {73.4} & {79.1} & {71.4}\\
    \midrule
    Lightweight decoder $h(g(I))$ & 86.5 & 79.6 & 88.5 & 82.0 & 80.7 & 72.4 & 71.5 & 63.0\\
    AutoSAM (ours) & {91.0} & \textbf{87.0} & {92.8} & \textbf{89.3} & \textbf{83.0} & \textbf{76.7} & \textbf{79.7} & \textbf{74.0}\\
    \bottomrule
    \end{tabular}
\end{center}
    \caption{Polyp Segmentation benchmarks results}
    \label{tab:polyp}
\end{table}

\begin{table*}[t!]
    \centering
    \footnotesize
    \renewcommand{\arraystretch}{1}
    \setlength\tabcolsep{1pt}
    \begin{center}
    \begin{adjustbox}{width=1\columnwidth}
    \label{tab:ModelScore}
    \begin{tabular}{c lcccccc cccccc} 
    \toprule
    
    & \multicolumn{1}{l}{\multirow{2}{*}{Method} }& \multicolumn{6}{c}{{SUN-SEG-\texttt{Easy}}} &\multicolumn{6}{c}{{SUN-SEG-\texttt{Hard}}}\\
   \cmidrule(rl){3-8}
   \cmidrule(rl){9-14}
    &  & $\mathcal{S}_{\alpha}$ & $E_\phi^{mn}$ & $F_\beta^w$ & $F_\beta^{mn}$  & Dice & Sen
    & $\mathcal{S}_{\alpha}$ & $E_\phi^{mn}$ & $F_\beta^w$ & $F_\beta^{mn}$  & Dice & Sen\\
    \midrule 
    \multirow{6}{*}{\rotatebox[origin=C]{90}{\textbf{Image-based}}}
    & UNet~\cite{ronneberger2015u} & 0.669 & 0.677 & 0.459 & 0.528 & 0.530 & 0.420 & 0.670 & 0.679 & 0.457 & 0.527 & 0.542 & 0.429 \\
    & UNet++~\cite{zhou2018unetplus} & 0.684 & 0.687 & 0.491 & 0.553 & 0.559 & 0.457 & 0.685 & 0.697 & 0.480 & 0.544 & 0.554 & 0.467 \\
    & ACSNet~\cite{zhang2020adaptive} & 0.782 & 0.779 & 0.642 & 0.688 & 0.713 & 0.601 & 0.783 & 0.787 & 0.636 & 0.684 & 0.708 & 0.618 \\
    & PraNet~\cite{fan2020pra} & 0.733 & 0.753 & 0.572 & 0.632 & 0.621 & 0.524 & 0.717 & 0.735 & 0.544 & 0.607 & 0.598 & 0.512 \\
    & SANet~\cite{wei2021shallow} & 0.720 & 0.745 & 0.566 & 0.634 & 0.649 & 0.521 & 0.706 & 0.743 & 0.526 & 0.580 & 0.598 & 0.505 \\
        & AutoSAM(ours) & \textbf{0.815} & \textbf{0.855} & \textbf{0.716} & \textbf{0.774} & 0.753 & \textbf{0.672}& \textbf{0.822} & \textbf{0.866} & \textbf{0.714} & \textbf{0.764} & \textbf{0.759} & \textbf{0.726} \\
    \midrule
    \multirow{9}{*}{\rotatebox[origin=C]{90}{\textbf{Video-based}}} 
    & COSNet~\cite{lu2019see} & 0.654 & 0.600 & 0.431 & 0.496 & 0.596 & 0.359 & 0.670 & 0.627 & 0.443 & 0.506 & 0.606 & 0.380 \\
    & MAT~\cite{zhou2020matnet} & 0.770 & 0.737 & 0.575 & 0.641 & 0.710 & 0.542 & 0.785 & 0.755 & 0.578 & 0.645 & 0.712 & 0.579 \\
    & PCSA~\cite{gu2020pyramid} & 0.680 & 0.660 & 0.451 & 0.519 & 0.592 & 0.398 & 0.682 & 0.660 & 0.442 & 0.510 & 0.584 & 0.415 \\
    & 2/3D~\cite{puyal2020endoscopic} & 0.786 & 0.777 & 0.652 & 0.708 & 0.722 & 0.603 & 0.786 & 0.775 & 0.634 & 0.688 & 0.706 & 0.607 \\
    & AMD~\cite{liu2021emergence} & 0.474 & 0.533 & 0.133 & 0.146 & 0.266 & 0.222 & 0.472 & 0.527 & 0.128 & 0.141 & 0.252 & 0.213 \\
    & DCF~\cite{zhang2021dynamic} & 0.523 & 0.514 & 0.270 & 0.312 & 0.325 & 0.340 & 0.514 & 0.522 & 0.263 & 0.303 & 0.317 & 0.364 \\
    & FSNet~\cite{ji2021full} & 0.725 & 0.695 & 0.551 & 0.630 & 0.702 & 0.493 & 0.724 & 0.694 & 0.541 & 0.611 & 0.699 & 0.491 \\
    & PNSNet~\cite{ji2021pnsnet} & 0.767 & 0.744 & 0.616 & 0.664 & 0.676 & 0.574 & 0.767 & 0.755 & 0.609 & 0.656 & 0.675 & 0.579 \\
    & VPS+~\cite{ji2022video} & 0.806 & 0.798 & 0.676 & 0.730 & \textbf{0.756} & 0.630& 0.797 & 0.793 & 0.653 & 0.709 & 0.737 & 0.623 \\
    \bottomrule
    \end{tabular}
    \end{adjustbox}
    \end{center}

    \caption{Quantitative results of two test sub-datasets from the SUN-SEG~\cite{ji2022video} dataset. Although being image-based, our method competes with video-based approaches, achieving SOTA performance in almost every benchmark. The best values are highlighted in \textbf{bold}.} 
    \label{tab:video}

\end{table*}

\paragraph{Results} The MoNu dataset results are reported in Tab.~\ref{tab:Monu}. We outperform all baselines, {including the latest Axial attention Unet \cite{wang2020axial}, Medical transformer \cite{valanarasu2021medical}, 3P-Seg~\cite{shaharabany2022end} and MedAdaptorSAM~\cite{wu2023medical}} , for both the Dice score and Mean-IoU. Our algorithm also performs better than the fully convolutional segmentation network with the same backbone Hardnet-85 (3P-Seg also uses the same backbone). Sample results for this and other datasets are presented in  Fig.~\ref{fig:Monu_seg}. 

The results for GlaS are also shown in Tab.~\ref{tab:Monu}. Our algorithm outperforms the Medical transformer by almost 10\% IoU \cite{shaharabany2022end}, 3P-SEG by almost 3\%, and MedAdaptor-SAM by more than a percent, despite the latter utilizing additional information in the form of ground truth points that are placed on the desired objects. 
Fig.~\ref{fig:MedAdapter} shows a visual comparison between our solution for SAM, with another MedAdaptor-SAM~\cite{wu2023medical}.

We also compare our algorithm using different types of prompts with the original prompt encoders of SAM. Tab.~\ref{tab:Monu} shows that our solution for medical prompts improves dramatically the performance of SAM, without any fine-tuning for SAM. Fig.~\ref{fig:Monu_seg} illustrates the gap in the accuracy between our solution and the one that uses the original prompt encoders of SAM.

It is intriguing that SAM encounters difficulties segmenting accurately medical images despite the availability of various prompts, including those based on ground truth. Nevertheless, as our method demonstrates, SAM is capable of delivering state-of-the-art segmentation outcomes without altering the core encoder and decoder modules for the learned prompts.

This phenomenon may have originated from two possible causes. One potential factor is the precision at which SAM incorporates the information encoded by $g$. Alternatively, a latent signal, analogous to adversarial noise, could be present, which alters the classification of the image without causing significant changes in its appearance.

The results for the Polyp datasets are listed in Tab.~\ref{tab:polyp}. In terms of IOU metric, our method outperforms the state-of-the-art on this benchmark 3P-SEG~\cite{shaharabany2022end} and Polyp-PVT~\cite{dong2021polyp}. For all the four dataset Kvasir-SEG and ClinicDB, ColonDB, and ETIS our algorithm achieved state-of-the-art results with a gap of 0.5, 0.3, 3.3 and 2.6 respectively. With respect to the DICE metric, our method outperforms other methods in two out of four datasets. 

The results for the SUN-SEG video dataset are listed in Tab.~\ref{tab:video}. For the SUN-SEG-Hard (unseen) dataset our method outperforms the state-of-the-art~\cite{ji2022video} on every metric tested, i.e. $\mathcal{S}_{\alpha}$ , $E_\phi^{mean}$ , $F_\beta^w$ , $F_\beta^{mean}$ , Dice \& Sen with a margin of 2.5, 7.3, 6.1, 5.5, 2.2, 10.3 respectively. For the SUN-SEG-Easy (unseen) dataset our method outperforms State-Of-The-Art methods in every metric except for the Dice-Score, which achieves 0.3 below VPS+~\cite{ji2022video}. Note that we are using an image-based method and outperforming the video-based methods that significantly outperform any other image-based method.

Finally, we measure the performances of the lightweight decoder $h$ for all the medical image datasets. As can be seen in Tab.~\ref{tab:Monu} and 
Tab.~\ref{tab:polyp}, $h(g(I))$ achieves a reasonable mask, although not as good as the output of SAM with $g$ prompts. A visual comparison of $h(g(I))$ and AutoSAM on the same $g(I)$ is shown in Fig.~\ref{fig:inverting}.

\section{Conclusions}

SAM is a powerful segmentation model for natural images. It has the potential to become a prominent foundation model, i.e., be effective for downstream tasks such as medical image analysis. We show that this may only require ``the right guidance'' in the form of a dedicated conditioning signal that is provided by an auxiliary network $g$ that replaces the prompt embedding. As no prompt is required, our method turns SAM into a fully automatic method.

In future work, we plan to learn one $g$ network for multiple medical imaging domains. It would be interesting to learn how well this ``universal-AutoSAM'' generalizes to new tasks without further training.

\bibliographystyle{splncs04}
\bibliography{medical}

\end{document}